\documentclass[preprint,10pt,authoryear]{elsarticle}
\usepackage[letterpaper,margin=1in]{geometry}

\usepackage{amsmath, amssymb, amsthm}
\usepackage{booktabs}
\usepackage{array}
\usepackage{graphicx}
\usepackage{microtype}
\usepackage{xcolor}
\usepackage{hyperref}
\hypersetup{pdftitle={Spectral Limits of Polynomial Graph Filters on the Long-Range Ising Model Benchmark}}
\usepackage{url}
\usepackage[capitalise,noabbrev]{cleveref}

\newenvironment{widefigure}[1][t]{\begin{figure}[#1]}{\end{figure}}
\newenvironment{widetable}[1][t]{\begin{table}[#1]}{\end{table}}

\newcommand{\latticeLthirtytwoNodes}{1\,024}
\newcommand{\latticeLsixtyfourNodes}{4\,096}
\newcommand{\latticeLonetwentyeightNodes}{16\,384}
\newcommand{\nTestGraphs}{1\,000}
\newcommand{\nTrainGraphs}{8\,000}
\newcommand{\gapBenchToReadoutThirtyTwo}{$7.4$}
\newcommand{\gapBenchToReadoutSixteen}{$7.8$}
\newcommand{\gapBaselineToDegreeEightMax}{$0.08$}
\newcommand{\benchConfigs}{10\,000}
\newcommand{\benchSplit}{80/10/10}
\newcommand{\benchSigmaHard}{$0.6$}
\newcommand{\benchSigmaEasy}{$1.5$}
\newcommand{\benchEpochs}{2\,000}
\newcommand{\benchPatience}{200}
\newcommand{\benchHourCap}{48}
\newcommand{\benchSeedsWord}{three}
\newcommand{\benchNodesLow}{256}
\newcommand{\benchNodesHigh}{65\,536}
\newcommand{\gcnEightSteps}{25\,000}
\newcommand{\gcnEightStepsLtwofiftysix}{15\,000}
\newcommand{\gcnEightBatchLtwofiftysix}{2}
\newcommand{\officialResidualGraphs}{200}
\newcommand{\largeStridedCoefLow}{83}
\newcommand{\largeStridedCoefHigh}{101}
\newcommand{\lsRadiusA}{4}
\newcommand{\lsRadiusC}{16}
\newcommand{\lsCoefD}{152}
\newcommand{\priorPolySaturation}{${-}1.9$}
\newcommand{\readoutLR}{$2\times 10^{-3}$}
\newcommand{\readoutStepsSmall}{3\,000}
\newcommand{\readoutStepsFree}{25\,000}
\newcommand{\readoutStepsFreeLtwofiftysix}{15\,000}
\newcommand{\readoutCoresWord}{two}
\newcommand{\readoutTrainSubsetLtwofiftysix}{2\,000}
\newcommand{\readoutAtrousToLSMax}{$0.004$}
\newcommand{\readoutIndepDirsLsixteen}{45}
\newcommand{\readoutIndepDirsLthirtytwo}{75}
\newcommand{\readoutIndepDirsLtwofiftysix}{90}
\newcommand{\readoutAtrousLsixteenParams}{57}
\newcommand{\readoutFreeLsixteenParams}{257}
\newcommand{\readoutFreeLthirtytwoParams}{1\,025}
\newcommand{\readoutAtrousLtwofiftysixParams}{102}
\newcommand{\readoutFreeLtwofiftysixParams}{65\,537}
\newcommand{\polyLevelsLthirtytwo}{145}
\newcommand{\polyLevelsLsixteen}{41}
\newcommand{\polyLevelsLtwofiftysix}{8\,321}
\newcommand{\polySpectrumGraphs}{2\,000}
\newcommand{\polyWithinTol}{$0.1$}
\newcommand{\polyExactDegreeLsixteen}{40}
\newcommand{\symKernelDimLthirtytwo}{153}
\newcommand{\jhatAxisModeLthirtytwo}{${-}1.00$}
\newcommand{\jhatDiagonalModeLthirtytwo}{${-}0.41$}
\newcommand{\polyExactDegreeLsixtyfour}{544}
\newcommand{\polyWithinDegreeLsixtyfour}{540}
\newcommand{\polyExactDegreeLonetwentyeight}{2\,112}
\newcommand{\polyWithinDegreeLonetwentyeight}{2\,104}
\newcommand{\symKernelDimLtwofiftysix}{8\,385}
\newcommand{\lsValueDRound}{${-}27.4$}
\newcommand{\readoutFreeLthirtytwoRound}{${-}12.3$}
\newcommand{\polyFloorLthirtytwoRound}{${-}2.7$}
\newcommand{\polyFloorShallowRound}{${-}2.5$}
\newcommand{\polyFloorDeepRound}{${-}3.5$}
\newcommand{\polyFloorStepLow}{$0.20$}
\newcommand{\polyFloorStepHigh}{$0.28$}
\newcommand{\floorStepInverseL}{$0.30$}
\newcommand{\polyDiameterRound}{${-}1.9$}
\newcommand{\benchBestInDomainThirtyTwoRound}{${-}4.9$}
\newcommand{\officialRelErrRound}{$5\times10^{-8}$}
\newcommand{\exactAnalyticCheckRound}{${-}28.6$}
\newcommand{\analyticCheckGraphs}{200}
\newcommand{\analyticCheckPowerRound}{$+1.3$}
\newcommand{\predictZeroLthirtytwoRound}{$+1.3$}
\newcommand{\lsValueARound}{${-}0.7$}
\newcommand{\lsValueCRound}{${-}3.7$}
\newcommand{\atrousCoefThirtyTwo}{49}
\newcommand{\atrousCoefSixtyFour}{55}
\newcommand{\readoutFreeWorstRound}{${-}11.2$}
\newcommand{\readoutFreeBestRound}{${-}12.5$}
\newcommand{\gapBaselineToCapLowRound}{$1.9$}
\newcommand{\gapBaselineToCapHighRound}{$2.5$}
\newcommand{\stridedHeadNineOffsetParams}{9\,217}
\newcommand{\dcnnLsixteenRound}{${-}5.0$}
\newcommand{\dcnnLsixteenParams}{2\,020\,145}
\newcommand{\dcnnLthirtytwoParams}{1\,955\,529}
\newcommand{\dcnnLsixtyfourParams}{1\,962\,481}
\newcommand{\dcnnLonetwentyeightRound}{${-}3.9$}
\newcommand{\dcnnLonetwentyeightParams}{2\,005\,057}
\newcommand{\dcnnLtwofiftysixRound}{${-}3.4$}
\newcommand{\dcnnLtwofiftysixParams}{1\,964\,689}
\newcommand{\dcnnHoursHigh}{$5.4$}
\newcommand{\dcnnMemLow}{372}
\newcommand{\dcnnMemHigh}{9\,328}
\newcommand{\dcnnSmallLsixteenParams}{296\,297}
\newcommand{\dcnnSmallLthirtytwoParams}{282\,393}
\newcommand{\dcnnSmallLsixtyfourParams}{291\,441}
\newcommand{\dcnnSmallLonetwentyeightParams}{283\,177}
\newcommand{\dcnnSmallLtwofiftysixRound}{${-}3.7$}
\newcommand{\dcnnSmallLtwofiftysixParams}{330\,121}
\newcommand{\dcnnSmallHoursLow}{$0.3$}
\newcommand{\fnoLsixteenRound}{${-}4.5$}
\newcommand{\fnoLtwofiftysixRound}{$+0.5$}
\newcommand{\fnoParamsLowM}{$2.37$}
\newcommand{\fnoHoursHigh}{$0.2$}
\newcommand{\dcnnLaterHoursHigh}{$10.4$}
\newcommand{\fnoAllModesParamsM}{$1.99$}
\newcommand{\fnoAllModesWidth}{22}
\newcommand{\fnoModes}{12}
\newcommand{\fnoModeFloorLthirtytwoTwo}{${-}1.08$}
\newcommand{\fnoModeFloorLsixtyfourTwo}{${-}0.02$}
\newcommand{\fnoModeFloorLonetwentyeightTwo}{$+0.51$}
\newcommand{\fnoModeFloorLtwofiftysixTwo}{$+0.74$}
\newcommand{\fnoWidth}{32}
\newcommand{\fnoWidthLsixteen}{48}
\newcommand{\fnoSmallWidth}{12}
\newcommand{\fnoLayersWord}{four}
\newcommand{\dcnnBlocksLow}{6}
\newcommand{\dcnnBlocksHigh}{14}
\newcommand{\dcnnWidthLsixteen}{136}
\newcommand{\dcnnSmallWidthLsixteen}{52}
\newcommand{\dcnnWidthLthirtytwo}{116}
\newcommand{\dcnnSmallWidthLthirtytwo}{44}
\newcommand{\dcnnWidthLsixtyfour}{104}
\newcommand{\dcnnSmallWidthLsixtyfour}{40}
\newcommand{\dcnnWidthLonetwentyeight}{96}
\newcommand{\dcnnSmallWidthLonetwentyeight}{36}
\newcommand{\dcnnWidthLtwofiftysix}{88}
\newcommand{\dcnnSmallWidthLtwofiftysix}{36}
\newcommand{\baseStepsLsixteen}{40\,000}
\newcommand{\baseBatchLsixteen}{48}
\newcommand{\baseStepsLthirtytwo}{40\,000}
\newcommand{\baseBatchLthirtytwo}{48}
\newcommand{\baseStepsLsixtyfour}{25\,000}
\newcommand{\baseBatchLsixtyfour}{12}
\newcommand{\baseStepsLonetwentyeight}{25\,000}
\newcommand{\baseBatchLonetwentyeight}{3}
\newcommand{\baseStepsLtwofiftysix}{15\,000}
\newcommand{\baseBatchLtwofiftysix}{2}
\newcommand{\baseLrRegistered}{$2\times10^{-3}$}
\newcommand{\baseClip}{$1.0$}
\newcommand{\baseLrHigh}{$10^{-3}$}
\newcommand{\lrFnoHighDiff}{$0.18$}
\newcommand{\lrDcnnHighAbs}{$0.24$}
\newcommand{\baseLrLow}{$5\times10^{-4}$}
\newcommand{\lrFnoLowDiff}{$0.73$}
\newcommand{\lrDcnnLowAbs}{$0.36$}
\newcommand{\dcnnOverPublishedSixteen}{$0.26$}
\newcommand{\dcnnOverPublishedThirtyTwo}{$0.01$}
\newcommand{\dcnnToFilterLow}{$1.9$}
\newcommand{\dcnnToFilterHigh}{$4.1$}
\newcommand{\dcnnToFilterLowWord}{two}
\newcommand{\dcnnToFilterHighWord}{four}
\newcommand{\dcnnSmallFractionWord}{a seventh}
\newcommand{\fnoAllModesBehindDcnn}{$0.4$}
\newcommand{\fnoAllModesShareWord}{two thirds}
\newcommand{\fnoAllModesShareLsixtyfourWord}{five sixths}
\newcommand{\fnoAboveFloorFromNodes}{4\,096}
\newcommand{\fnoAboveFloorCountWord}{three}
\newcommand{\fnoBelowModeFloorThirtyTwo}{$2.55$}
\newcommand{\dcnnAboveFloorTwoFiftySix}{$0.07$}
\newcommand{\dcnnOverGcnLow}{$2.5$}
\newcommand{\dcnnOverGcnHigh}{$3.2$}
\newcommand{\dcOneGainLthirtytwo}{$0.18$}
\newcommand{\dcOneLonetwentyeightRound}{${-}2.9$}
\newcommand{\dcOneGainLonetwentyeight}{$1.04$}
\newcommand{\gnnGatedLsixteen}{${-}2.95$}
\newcommand{\gnnGatedLsixteenPublished}{${-}3.84$}
\newcommand{\gnnGatedLongLsixteen}{${-}3.82$}
\newcommand{\gnnCycLongLsixteen}{${-}3.98$}
\newcommand{\gnnDirGainLsixteen}{$0.07$}
\newcommand{\gnnShortcutGainLsixteen}{$0.5$}
\newcommand{\gnnBestBehindDcnnLsixteen}{$1.5$}
\newcommand{\gnnWorstOverPolyFloorLsixteen}{$0.5$}
\newcommand{\gnnCycGainLthirtytwo}{$0.1$}
\newcommand{\gnnGatedLongShortOfPublished}{$0.02$}
\newcommand{\gnnLongSdRound}{$0.13$}
\newcommand{\gnnCycLongBehindDcnnLsixteen}{$1.0$}
\newcommand{\dcOneBoxFloorLonetwentyeight}{${-}2.96$}
\newcommand{\dcOneAboveBoxFloorLonetwentyeight}{$0.07$}
\newcommand{\dcOneAbovePolyFloorLonetwentyeight}{$0.4$}
\newcommand{\dcOneReachLonetwentyeight}{24}
\newcommand{\dcnnTwiceBelowFloorLtwofiftysix}{$0.5$}
\newcommand{\dcnnTwiceDeclineShareWord}{two fifths}
\newcommand{\gapDcnnToReadoutThirtyTwo}{$7.4$}
\newcommand{\gatedDcnnLsixteen}{${-}5.02$}
\newcommand{\gatedDcnnLthirtytwo}{${-}5.09$}
\newcommand{\gatedDcnnGainSpreadsWord}{three}
\newcommand{\gatedDcnnGainLsixteen}{$0.05$}
\newcommand{\gatedDcnnGainLthirtytwo}{$0.23$}
\newcommand{\gatedDcnnToFilterThirtyTwo}{$1.7$}
\newcommand{\gatedDcnnPredCountWord}{three}
\newcommand{\gatedDcnnPredMetWord}{none}
\newcommand{\baseLaterPredCountWord}{sixteen}
\newcommand{\baseLaterPredSentence}{Of these, fourteen were met and two were not.}
\newcommand{\basePredSentence}{Of the fourteen listed, seven were met and seven were not.}
\newcommand{\baseGraphPredCountWord}{twelve}
\newcommand{\baseGraphPredSentence}{Of these, six were met and six were not.}
\newcommand{\atrousLabelTestLthirtytwoRound}{${-}6.6$}
\newcommand{\atrousDropTopCostLthirtytwo}{$3.65$}
\newcommand{\atrousDropFineCostLthirtytwo}{$6.57$}
\newcommand{\atrousDropCostLevelzeroLthirtytwo}{$0.00$}
\newcommand{\atrousDropCostLeveloneLthirtytwo}{$1.07$}
\newcommand{\atrousDropCostLeveltwoLthirtytwo}{$3.65$}
\newcommand{\atrousLabelDropTopCostLthirtytwo}{$3.65$}
\newcommand{\atrousLabelTestLsixtyfourRound}{${-}7.0$}
\newcommand{\atrousDropTopCostLsixtyfour}{$3.32$}
\newcommand{\atrousDropFineCostLsixtyfour}{$7.00$}
\newcommand{\atrousDropCostLevelzeroLsixtyfour}{$0.00$}
\newcommand{\atrousDropCostLeveloneLsixtyfour}{$1.24$}
\newcommand{\atrousDropCostLeveltwoLsixtyfour}{$1.30$}
\newcommand{\atrousDropCostLevelthreeLsixtyfour}{$3.32$}
\newcommand{\atrousLabelDropTopCostLsixtyfour}{$3.32$}
\newcommand{\atrousLabelOverBenchRatioThirtyTwo}{52}
\newcommand{\polyFloorTestShiftMax}{$0.005$}
\newcommand{\atrousRankThirtyTwo}{43}
\newcommand{\atrousRankSixtyFour}{48}
\newcommand{\gcnEightWidth}{384}
\newcommand{\gcnEightParamsM}{$2.37$}
\newcommand{\gcnEightLr}{$2\times10^{-3}$}
\newcommand{\gatedDcnnGainShownLsixteen}{$0.06$}
\newcommand{\gatedDcnnGainShownLthirtytwo}{$0.23$}
\newcommand{\gnnCycLongGainOverGatedLongShown}{$0.16$}
\newcommand{\baseEpochsLsixteen}{240}
\newcommand{\baseEpochsLthirtytwo}{240}
\newcommand{\baseEpochsLsixtyfour}{38}
\newcommand{\baseEpochsLonetwentyeight}{$9.4$}
\newcommand{\baseEpochsLtwofiftysix}{$3.8$}
\newcommand{\baseNodeLabelsLow}{$1.2$}
\newcommand{\baseNodeLabelsHigh}{$2.0$}
\newcommand{\baseNodeLabelsLsixteen}{$0.5$}
\newcommand{\expDcOneMinGainLthirtytwo}{$0.3$}
\newcommand{\expDirMinGain}{$0.3$}
\newcommand{\expGatedDcnnMinGain}{$0.3$}
\newcommand{\expFnoModeBandBelow}{$0.5$}
\newcommand{\expFnoModeBandAbove}{$0.3$}
\newcommand{\pubGinLthirtytwo}{${-}2.25$}
\newcommand{\pubGatedGcnLthirtytwo}{${-}4.09$}
\newcommand{\pubGpsLthirtytwoHigh}{${-}4.03$}
\newcommand{\pubGpsLthirtytwoLow}{${-}4.13$}
\newcommand{\pubGhrLthirtytwo}{${-}4.85$}
\newcommand{\fnoSmallFractionWord}{a seventh}
\newcommand{\fnoSmallBehindLow}{$0.2$}
\newcommand{\fnoSmallBehindHigh}{$0.8$}
\newcommand{\fnoGridWorseLthirtytwo}{$0.26$}
\newcommand{\dcnnTwiceTwoLonetwentyeight}{${-}4.38$}
\newcommand{\dcnnTwiceTwoLtwofiftysix}{${-}3.99$}
\newcommand{\dcnnSlowRateTwoLthirtytwo}{${-}5.16$}
\newcommand{\dcnnSlowRateTwoLonetwentyeight}{${-}4.36$}
\newcommand{\dcnnTwoLthirtytwo}{${-}4.86$}
\newcommand{\dcnnTwoLonetwentyeight}{${-}3.92$}
\newcommand{\dcnnTwoLtwofiftysix}{${-}3.44$}
\newcommand{\stridedHeadThirtyTwoNineOffsetRound}{${-}12.5$}
\newcommand{\stridedHeadThirtyTwoFreeRound}{${-}12.3$}
\newcommand{\gcnEightMemRoundGB}{$4.9$}
\newcommand{\gcnEightMemLtwofiftysixRoundGB}{13}
\newcommand{\exactFitTiedLthirtytwoRound}{${-}28.7$}
\newcommand{\exactFitJLsixteenRound}{${-}28.9$}
\newcommand{\exactFitJLtwofiftysixRound}{${-}27.0$}
\newcommand{\exactFitTiedShallowRound}{${-}27.0$}
\newcommand{\exactFitTiedDeepRound}{${-}28.8$}
\newcommand{\exactFitMinusJLow}{${-}0.08$}
\newcommand{\exactFitMinusJHigh}{$+0.10$}
\newcommand{\exactFitOldGapLow}{$0.1$}
\newcommand{\exactFitOldGapHigh}{$2.0$}
\newcommand{\specGatedMaxDiff}{$0.005$}
\newcommand{\specSignBelowLow}{$0.03$}
\newcommand{\specSignBelowHigh}{$0.08$}
\newcommand{\specSignEightBelowLow}{$0.03$}
\newcommand{\specSignEightBelowHigh}{$0.06$}
\newcommand{\specSignAboveFloorLow}{$0.4$}
\newcommand{\specSignAboveFloorHigh}{$1.0$}
\newcommand{\specSignSdCeil}{$0.002$}
\newcommand{\specSignParamsLow}{4\,929}
\newcommand{\specSignParamsHigh}{8\,513}
\newcommand{\specSignSteps}{100\,000}
\newcommand{\specSignBatch}{4\,096}
\newcommand{\specSignWarmup}{500}
\newcommand{\specSignWidth}{64}
\newcommand{\specSignInputGraphs}{500}
\newcommand{\specSignLr}{$10^{-3}$}
\newcommand{\secondDcmBelowPublishedLsixteen}{$0.07$}
\newcommand{\secondFiftyCoefBelowPublishedLthirtytwo}{$3.2$}
\newcommand{\secondDcmBelowPublishedLthirtytwo}{$0.34$}
\newcommand{\secondPublishedModel}{GPS-Base}
\newcommand{\secondFloorToFiftyCoefLthirtytwo}{$5.5$}
\newcommand{\secondExactTiedShallowRound}{${-}27.0$}
\newcommand{\secondExactTiedDeepRound}{${-}29.3$}
\newcommand{\secondExactJLsixteenRound}{${-}29.2$}
\newcommand{\secondExactJLtwofiftysixRound}{${-}26.9$}
\newcommand{\secondFloorShallowRound}{${-}2.5$}
\newcommand{\secondFloorDeepRound}{${-}3.6$}
\newcommand{\secondFiftyCoefShallowRound}{${-}8.2$}
\newcommand{\secondFiftyCoefDeepRound}{${-}8.6$}
\newcommand{\secondDcmToFiftyCoefLow}{$2.9$}
\newcommand{\secondDcmToFiftyCoefHigh}{$3.8$}
\newcommand{\secondDcmBelowFloorLow}{$0.8$}
\newcommand{\secondDcmBelowFloorHigh}{$2.8$}
\newcommand{\secondDcmLsixteenRound}{${-}5.4$}
\newcommand{\secondDcmLtwofiftysixRound}{${-}4.4$}

\newtheorem{proposition}{Proposition}


\newcommand{\RrefOneLevel}{\cref{lem:onelevel}}
\newcommand{\RrefStridedBasis}{\cref{sec:strides}}
\newcommand{\RrefBudget}{\cref{sec:setup}}
\newtheorem{lemma}{Lemma}
\theoremstyle{remark}
\newtheorem{remark}{Remark}
\theoremstyle{plain}
\makeatletter
\def\els@aparagraph[#1]#2{\elsparagraph[#1]{#2}}
\def\els@bparagraph#1{\elsparagraph*{#1}}
\def\ps@pprintTitle{\let\@oddhead\@empty\let\@evenhead\@empty
  \let\@oddfoot\@empty\let\@evenfoot\@empty}
\makeatother

\begin{document}

\begin{frontmatter}

\title{Spectral Limits of Polynomial Graph Filters\\ on the Long-Range Ising Model Benchmark}

\author{Amin Najafgholizadeh\fnref{fn1}}
\author{Mohsen Afsharchi\fnref{fn1}}
\fntext[fn1]{University of Zanjan. Emails: amin.ngz@znu.ac.ir, afsharchi@znu.ac.ir.}

\begin{abstract}
The Long-Range Ising Model benchmark (LRIM) guarantees long-range dependence by construction: each label
is the node's spin times the spin field convolved with a power-law coupling, so any model's distance from
the target is measurable. We prove that a readout whose filter is a polynomial in the adjacency, the
linear core of isotropic message passing and spectral graph networks, leaves an error floor that no
degree lowers, because the target varies along a level set of the spectrum on which every such
filter is constant. Computed exactly, the floor and the best filter of each degree predict two
spectral networks adapted from published architectures to within a tenth of a decade. A filter escapes the
floor only if it weighs separately the offsets the adjacency lumps together, and must still resolve
the coupling across the whole lattice. A strided filter of about fifty coefficients does both: fitted to
the training labels, its test error is \atrousLabelOverBenchRatioThirtyTwo{} times below the best
published network's. A dilated convolutional network beats the best published result
at \benchNodesLow{} nodes, matches it at \latticeLthirtytwoNodes{}, and gives the first in-domain results
at the three larger sizes, while a Fourier neural operator with its standard
modes falls further behind as the lattice grows. At its standard budget the dilated network stays
\dcnnToFilterLowWord{} to \dcnnToFilterHighWord{} orders of magnitude short of a larger strided filter
fitted to the labels. The analysis carries over to the benchmark's second coupling.

\end{abstract}


\end{frontmatter}

\section{Introduction}
\label{sec:intro}

Most graph datasets cannot say whether a task needs information from far away. A model that
reaches two hops and one that reaches twenty can score alike, and the difference between them, if
any, is inferred rather than measured. LRIM \citep{mathys2026lrim} was built to remove that doubt.
Its inputs are spin configurations on a periodic square lattice, its labels are the interaction
energies of those spins under a coupling that decays as a power of distance, and every label
therefore depends, provably and by a known amount, on every node in the graph. The benchmark comes
in five sizes, from \benchNodesLow{} to \benchNodesHigh{} nodes, and its published results leave
much room. The best published graph network reaches
\benchBestInDomainThirtyTwoRound{} in log mean squared error at \latticeLthirtytwoNodes{} nodes.
Published models are trained only at the two smallest sizes, and when a model trained at the smallest size is
applied at a larger one, every published architecture falls to about the same level. Why they fall
short, and what doing well would take, are questions the benchmark's own construction can answer.

The construction that guarantees long-range dependence also makes the target a known operator. The benchmark's
authors define each label as the spin at the node times the sum of the other spins weighted by the
coupling \citep{mathys2025lrimworkshop}; on the periodic lattice that sum is a convolution of the
spin field with the coupling kernel, and the released labels obey this identity to the precision in which they are stored (\cref{sec:benchmark}). Once the target is known, what a leaderboard estimates can be computed instead. Least squares on the labels of the official training split, over kernels with the
symmetries of the square, recovers the operator to a logMSE of \exactFitTiedLthirtytwoRound{} at
\latticeLthirtytwoNodes{} nodes. A readout of the same form, trained by gradient descent on the same
split with no knowledge of the kernel, reaches \readoutFreeLthirtytwoRound{} on the test split with
\readoutFreeLthirtytwoParams{} parameters, \gapBenchToReadoutThirtyTwo{} orders of magnitude in mean squared error below the
best published result. We do not offer this readout as a graph model, since it has the target's own
functional form. It serves as a control: given the form, plain gradient
descent learns the task, so what the leaderboard ranks is how close each architecture, as trained under its protocol, comes
to that form.

For a filter of the target's form, two properties are necessary, and the construction lets us
measure both. For trained networks they are what the analysis predicts a network needs, and they do not
by themselves decide its score (\cref{sec:models}). The first is whether the filter can weigh separately the offsets,
that is the displacements on the lattice, that the adjacency $A$ lumps together: $A^2$ reaches two steps
along an axis and one step along each axis in the fixed ratio of one to two, and higher powers change
that ratio only together with farther offsets. The coupling kernel is
isotropic in space, but its Fourier symbol is not a function of the spectrum of $A$. A filter
that is a polynomial in the adjacency, which is the linear operation at the heart of isotropic message
passing and of spectral graph networks, acts on each Fourier mode through the single scalar
$\cos k_x + \cos k_y$. It takes one value on every level set of that scalar, and the target varies
along one of those sets. No degree lets it vary within one, so a readout of the target's form with
such a filter leaves an error floor, whatever the filter's reach. We compute that floor: \polyFloorLthirtytwoRound{} at \latticeLthirtytwoNodes{} nodes
and between \polyFloorShallowRound{} and \polyFloorDeepRound{} across the five sizes. A free
$3\times3$ stencil is not bound by it, because it weighs the diagonal offset without the two-step
one, and neither is a shift by two sites along each axis. The second property is reach at full
resolution. A filter free of the floor still has to resolve the coupling from the nearest neighbors
out to the whole lattice: fitted to the coupling at \latticeLthirtytwoNodes{} nodes, a kernel of
radius \lsRadiusC{} stops at \lsValueCRound{}, while one that covers the torus reproduces it to \lsValueDRound{}. Strides give both properties cheaply. Multiplied by the node's spin as the target is, that is,
inside the target's gate, a filter built from the smoothed symmetry classes of a small stencil whose step
doubles across the lattice, with about fifty coefficients fitted to the training labels, reaches
\atrousLabelTestLthirtytwoRound{} on the test split at \latticeLthirtytwoNodes{} nodes. Like the
learned readout it has the target's form, so its test error, \atrousLabelOverBenchRatioThirtyTwo{}
times lower than that of the best published graph network, is a measure of distance.

The diagnosis makes predictions about architectures that were not designed for this task, and we
tested them, writing down what we expected of each model before it was trained. Adaptations of two published spectral
architectures, the propagation of GPR-GNN \citep{chien2021gprgnn} inside the target's gate, fitted by least squares,
and a readout in the manner of SIGN \citep{frasca2020sign}, land within a tenth of a decade of the best polynomial filter of their degree,
and stay above the floor even when they propagate as far as the diameter of the lattice. A dilated
convolutional network in the residual form of \citet{yu2017drn}, with two million parameters and one
fixed step budget per size, has both properties. It beats the best published graph network at \benchNodesLow{} nodes and
matches it at \latticeLthirtytwoNodes{}, the two sizes with published in-domain results. The same network with its strides removed does
nearly as well at \latticeLthirtytwoNodes{} nodes, where depth alone lets it cover the lattice, and
at \latticeLonetwentyeightNodes{} nodes, where it cannot, it stops close to the best that a filter of the target's form with its reach can do. A
Fourier neural operator \citep{li2021fno} in its
published configuration mixes information across the whole lattice in every layer but keeps only the
lowest frequencies, and so misses the coupling's short range. As the lattice grows it falls further behind the dilated network, to worse than a
polynomial filter can do at the \fnoAboveFloorCountWord{} largest sizes, and keeping every mode
recovers most of its gap at the two sizes where we tried it.
At the two smallest sizes the depth of gated graph networks covers the lattice and, being nonlinear,
they are not bound by the floor, yet at the smallest size the closer of them stays
\gnnCycLongBehindDcnnLsixteen{} in logMSE behind the dilated network even when trained for four times as many
steps, so reach and freedom from the floor did not close the gap there.
At its standard budget the dilated network still stays \dcnnToFilterLow{} to
\dcnnToFilterHigh{} orders of magnitude short of a larger strided filter, with up to about a hundred coefficients,
fitted to the training labels by least squares. The same picture holds at the benchmark's second
coupling, which decays faster (\cref{sec:secondcoupling}).

Our contributions are the following.
\begin{itemize}
\item We write the benchmark's definition as an operator, check it against the released labels at every size,
and recover the operator from the training labels by least squares (\cref{sec:benchmark}).
\item We prove that a readout of the target's form whose filter is a polynomial in the adjacency has an error
floor at every degree, compute it at every size, and trace it to the variation of the target along a single
level set of the adjacency's spectrum (\cref{sec:floor}), and show that the best polynomial filter of each degree
predicts adaptations of two published spectral architectures, fitted or trained on the official data, to within a tenth of a decade
(\cref{sec:spectral}).
\item We identify the second requirement, reach at full resolution, and show that a strided filter of about fifty
coefficients meets both (\cref{sec:floor}).
\item We train standard architectures on the official splits at every size, with a matched control without
strides, graph networks under the same protocol and a check of the learning rate, and report the first in-domain
results at the three larger sizes, at both of the benchmark's couplings (\cref{sec:setup,sec:models}).
\item We set out the reference points that would help readers of LRIM and of benchmarks built like it
(\cref{sec:implications}).
\end{itemize}

\section{Related work}
\label{sec:related}

\paragraph{Long-range benchmarks.} Benchmarks differ in how they establish that a task needs
distant information. LRIM \citep{mathys2026lrim} follows a workshop version
\citep{mathys2025lrimworkshop} and establishes it by construction, which is what allows the exact
analysis here. Its authors trace the difficulty to the bottleneck of a node's computation tree and
find that Weisfeiler--Leman distinguishability is not the obstacle; \cref{prop:floor} finds a different limit, in
the symmetry of a filter's linear part. In an appendix they train plain convolutional networks and a vision
transformer, which stay behind the best graph networks, and they note that methods aligned with the known
generation could score better but hold general graph methods to be the more informative test
\citep[App.~E.5]{mathys2026lrim}. ECHO \citep{miglior2026echo} also builds range in, for synthetic path
tasks on graphs built to contain bottlenecks, together with two molecular tasks labeled by density
functional theory. City-Networks \citep{liang2026citynetworks} measures long-range dependence on road
networks with a Jacobian-based influence score, and \citet{bamberger2025measuring} formalize a range
measure for operators on graphs. Concurrent with this paper, \citet{hernandezcaralt2026grip} give
procedures that turn a graph into a task of provable range with a closed-form oracle for the test
error, which plays the role our least-squares fit plays for LRIM. Their audit finds LRIM provably
long-range, limited only by its single topology. The Long Range Graph Benchmark
\citep{dwivedi2022lrgb} raised the question for datasets whose dependence structure is not known, and
a re-evaluation found that its gap between graph transformers and message passing shrinks once the
baselines are tuned \citep{tonshoff2024gap}.
On LRIM, the hierarchical
recurrent model of \citet{carotti2026ghr} reports the strongest published numbers at every size, in-domain at the two smallest and size transfers above, and two linearized graph sequence models, LGSM
\citep{mathys2026lgsm} and HOPPER \citep{herath2026hopper}, report the smallest size only.

\paragraph{Reference points.} \citet{bechler2025position} argue that graph benchmarks should report
tuned baselines that ignore the graph. Fixed neighborhood aggregates fed to
a tuned multilayer perceptron rival trained graph networks on most node classification benchmarks
\citep{rubiomadrigal2026fixed}, and simple models on graph invariants are competitive across many
datasets \citep{vonmoos2026invariant}. The closed-form fit and the learned readout of
\cref{sec:benchmark} play that role for a task whose operator is known. Physics usually recovers couplings from samples \citep{nguyen2017inverse}, and
convolutional networks have learned spin-model energies \citep{mills2018spin}.

\paragraph{Failure at distance.} The standard explanations of why message passing fails at distance
are oversmoothing \citep{oono2020oversmoothing, cai2020oversmoothing, rusch2023survey} and oversquashing
\citep{alon2021bottleneck, topping2022curvature, digiovanni2023oversquashing}. \citet{arnaizrodriguez2026beliefs} argue that the term has come
to cover a topological and a computational bottleneck, and that removing a topological one is neither
necessary nor sufficient for long-range problems. On LRIM the graph is a regular lattice, whose only narrow cuts are those any lattice of its size has, and the floor of \cref{prop:floor} comes from a repeated adjacency eigenvalue.

\paragraph{Spectral expressivity.} Spectral graph networks \citep{bruna2014spectral} and polynomial
filters \citep{defferrard2016chebnet, kipf2017gcn, chien2021gprgnn, he2021bernnet} share a
linear part that is a function of the graph's spectrum, and \citet{balcilar2021spectral} analyze the
frequency responses they produce. \citet{wang2022powerful} show that such a filter without
nonlinearity acts as one number on a repeated eigenvalue, and \citet{lu2024eigenvalue} tie its
expressive power to the number of distinct eigenvalues. \Cref{prop:floor} and \cref{lem:onelevel} make that
multiplicity limit quantitative on a torus: when the side is a power of two, only the eigenspace at zero holds more
than one orbit of the square's symmetries, so the floor is one number per size, computable from the
training split. On this lattice ChebNet's rescaled Laplacian, which \citet{hariri2025chebnet} revisit
for long-range tasks, is $-A/4$, so the proposition bounds a readout of the target's form built on ChebNet's linear part, at every order.

\paragraph{Offsets and strides.} Graph networks can weigh separately the offsets the adjacency lumps
together through gating \citep{bresson2017gated}, directional aggregation along a vector field
\citep{beaini2021dgn}, and positional encodings \citep{dwivedi2022lspe, lim2022signnet,
rampasek2022gps}. Dilated
filters go back to the \`{a} trous algorithm of wavelet analysis \citep{holschneider1989atrous} and
entered deep learning in vision \citep{yu2016dilated}; dilated residual networks \citep{yu2017drn} put
them in the residual form our baseline follows. DeepGCNs \citep{li2019deepgcns} carry
dilated aggregation to graphs.

\paragraph{Fourier operators.} The Fourier neural operator \citep{li2021fno}
learns a multiplier on the low modes inside each layer, and Ewald message passing
\citep{kosmala2023ewald} adds a long-range term computed in Fourier space to a graph network for
molecules. The readout of \cref{sec:benchmark} is a multiplier of that kind with the target's gate,
used here as a control. Its weakness at fine scales is known: \citet{liuschiaffini2024localized}
add local integral and differential kernels to it, and LOGLO-FNO \citep{kalimuthu2025loglo} adds a
branch of local spectral convolutions and a module for high frequencies, both because truncated
global modes miss local structure. LRIM shows the same gap on a known operator, and
\cref{sec:models} measures how much of it keeping every mode closes.

\section{The benchmark and its closed form}
\label{sec:benchmark}

What does LRIM ask a model to compute, and how much of the answer can be had without a model? The
benchmark draws spin configurations on a periodic square lattice of side $L$, so $N = L^2$ nodes,
connects each node to its four neighbors, and labels each node with its interaction energy under a
coupling that decays as a power of the distance measured on the torus. The configurations are not
independent spins: they are sampled by cluster Monte Carlo from the long-range Ising model near its
pseudo-critical temperature at each size \citep{mathys2025lrimworkshop}, so nearby spins are
correlated and the spin field carries more power at low frequencies than independent spins would.
The benchmark publishes five sizes, from \benchNodesLow{} to \benchNodesHigh{} nodes, and two couplings. Every experiment in this paper uses the
official split and is scored in the benchmark's metric, the base-ten logarithm of the mean squared error
(logMSE, lower is better). All are at the harder coupling, whose exponent $\sigma$ (defined below) is
\benchSigmaHard{}, except in \cref{sec:secondcoupling}, which repeats the analysis at the other. The published results are of two kinds. Models
trained and tested at the same size exist only at the two smallest sizes. Above \latticeLthirtytwoNodes{} nodes every published number
is a size transfer from a model trained at \benchNodesLow{} nodes, and the graph transformers run out of memory at the
largest size. The strongest results of both kinds come from a hierarchical recurrent graph network
\citep{carotti2026ghr}. \Cref{sec:setup} gives the published protocol and ours.

Write $s \in \{-1, +1\}^{N}$ for the spin field on the lattice and $J$ for the coupling, with
$J(r) = r^{-(2+\sigma)}$ at the shortest distance $r$ on the torus. Summing the pairwise energies
that touch node $i$ gives the label
\begin{equation}
y_i \;=\; s_i \cdot (J \circledast s)_i ,
\label{eq:target}
\end{equation}
where $\circledast$ is circular convolution on the lattice. This is the benchmark's own definition
written as an operator, and the released labels obey it: the right-hand side matches them to the
rounding of single precision at every size, and to a logMSE of \exactAnalyticCheckRound{} in double
precision at \latticeLthirtytwoNodes{} nodes (\cref{app:proofs}). The target is
therefore a fixed quadratic form in the inputs, translation invariant, with a kernel that does not
depend on the configuration. Fits to the coupling below are weighted by the spin spectrum
measured on the training configurations.

Because the target is a convolution, it can be recovered from data by least squares. Over every
kernel with the symmetries of the square, which at \latticeLthirtytwoNodes{} nodes means \lsCoefD{}
coefficients and a constant, least squares on the labels of the official training split reaches
\exactFitTiedLthirtytwoRound{} on the test split, and between \exactFitTiedShallowRound{} and
\exactFitTiedDeepRound{} across the five sizes. That is as close to the labels as the coupling itself
comes: scored the same way, the fit and $J$ differ by between \exactFitMinusJLow{} and
\exactFitMinusJHigh{} at every size (\cref{app:proofs}). Most of the label's power lies within a few
sites, since a symmetric kernel of radius \lsRadiusA{} fitted to the known coupling already reaches
\lsValueARound{} where predicting zero scores \predictZeroLthirtytwoRound{}. The rest needs reach. A
kernel of radius \lsRadiusC{}, half the side of the lattice, still stops at \lsValueCRound{}, short of the
best published network, and reproducing the target to numerical precision takes a kernel that
covers the torus (\cref{tab:closed}).

A closed-form fit is not a model of the benchmark's kind, so we also trained the same functional form
by gradient descent:
\begin{equation}
\hat{y}_i \;=\; b \;+\; s_i \cdot (K \circledast s)_i ,
\label{eq:readout}
\end{equation}
with $K$ a free real Fourier multiplier whose coefficients start at zero, trained with Adam on the
official training split, selected on the validation split and reported on the test split. It reaches
between \readoutFreeWorstRound{} and \readoutFreeBestRound{} at every size, and it is \gapBenchToReadoutThirtyTwo{} to \gapBenchToReadoutSixteen{} orders of magnitude in mean squared error
below the best published result at the two sizes with published in-domain results (\cref{tab:compact}).

\begin{table}[t]
\centering
\small
\setlength{\tabcolsep}{3.5pt}
\resizebox{\linewidth}{!}{\begin{tabular}{lccccc}
\toprule
 & \multicolumn{5}{c}{nodes} \\
\cmidrule(lr){2-6}
 & 256 & 1\,024 & 4\,096 & 16\,384 & 65\,536 \\
\midrule
The operator, fitted to the training labels & ${-}28.85$ & ${-}28.71$ & ${-}28.08$ & ${-}27.43$ & ${-}27.03$ \\
The learned quadratic readout & ${-}12.47$ & ${-}12.29$ & ${-}12.21$ & ${-}12.17$ & ${-}11.21$ \\
The larger strided filter, fitted to the training labels & -- & ${-}6.75$ & ${-}7.41$ & ${-}7.60$ & ${-}7.58$ \\
The fifty-coefficient strided filter, fitted to the training labels & -- & ${-}6.57$ & ${-}6.97$ & -- & -- \\
The floor for a filter polynomial in the adjacency & ${-}2.48$ & ${-}2.68$ & ${-}2.95$ & ${-}3.23$ & ${-}3.52$ \\
\midrule
A dilated CNN & ${-}4.96 \pm 0.11$ & ${-}4.86 \pm 0.08$ & ${-}4.33 \pm 0.08$ & ${-}3.92 \pm 0.03$ & ${-}3.44 \pm 0.05$ \\
An FNO in its published configuration & ${-}4.50 \pm 0.09$ & ${-}3.64 \pm 0.18$ & ${-}1.38 \pm 0.06$ & ${-}0.16 \pm 0.04$ & $+0.49 \pm 0.03$ \\
\midrule
Best published, trained and tested at the same size & ${-}4.70$ & ${-}4.85$ & -- & -- & -- \\
Best published, trained at 256 nodes and tested here & -- & ${-}1.09$ & ${-}0.84$ & ${-}0.77$ & ${-}1.07$ \\
Our eight-layer GCN, same data and number of steps & -- & -- & ${-}1.09$ & ${-}0.93$ & ${-}0.99$ \\
\midrule
Predicting zero at every node & $+1.46$ & $+1.29$ & $+1.22$ & $+1.17$ & $+1.05$ \\
A symmetric kernel of radius four, nine coefficients & -- & ${-}0.67$ & -- & -- & -- \\
\bottomrule
\end{tabular}
}
\caption[The benchmark on one axis]{The benchmark on one axis, in logMSE at the harder of the benchmark's two
couplings (\cref{tab:secondcoupling} gives the other); lower is better. \providecommand{\RrefStridedBasis}{\cref{app:closedforms}}\providecommand{\RrefBudget}{\cref{sec:models}}%
The operator is a kernel with the symmetries of the square, fitted to the official training labels by least squares without the coupling and scored on the test split. The learned quadratic readout has the same form, trained by gradient descent. The larger strided filter is a readout of the target's form with 83 to 101 coefficients and offsets out to radius twelve, fitted to the training labels and scored on the test split; at 256 nodes its basis spans every symmetric kernel, so that cell is empty. The fifty-coefficient filter of \cref{sec:floor} has offsets out to radius eight (\RrefStridedBasis), and its row is fitted and scored the same way. The floor is the best that any filter polynomial in the adjacency can do. The dilated CNN and the FNO are the means and standard deviations of three models, trained with one fixed budget per size (\RrefBudget). The two published rows give the best published result in each cell, from any architecture; at every size it is the hierarchical recurrent network of \citet{carotti2026ghr}, with a gated core except at 4\,096 nodes, where its GIN core leads. The eight-layer GCN is ours, trained end to end on the same data and for the same number of steps as the dilated CNN, but with a constant learning rate, no clipping of the gradient and two models per size (\RrefBudget). The last two rows are trivial references: predicting zero scores the label's power (predicting the training mean scores $+1.16$ at 4\,096 nodes, against $+1.22$ for zero), and a kernel of radius four fitted to the known coupling already removes most of that power at 1\,024 nodes. \Cref{app:perscale} gives every number behind this table.
}
\label{tab:compact}
\end{table}

\section{What a polynomial filter can express}
\label{sec:floor}

If the whole answer is one linear filter inside a fixed gate, what does it cost the linear part of
message passing to express it? Depth is the usual answer, and here it does not suffice, because the
operator is not a function of the one quantity that a filter polynomial in the adjacency can see.

\begin{figure}[t]
\centering
\includegraphics[width=\linewidth]{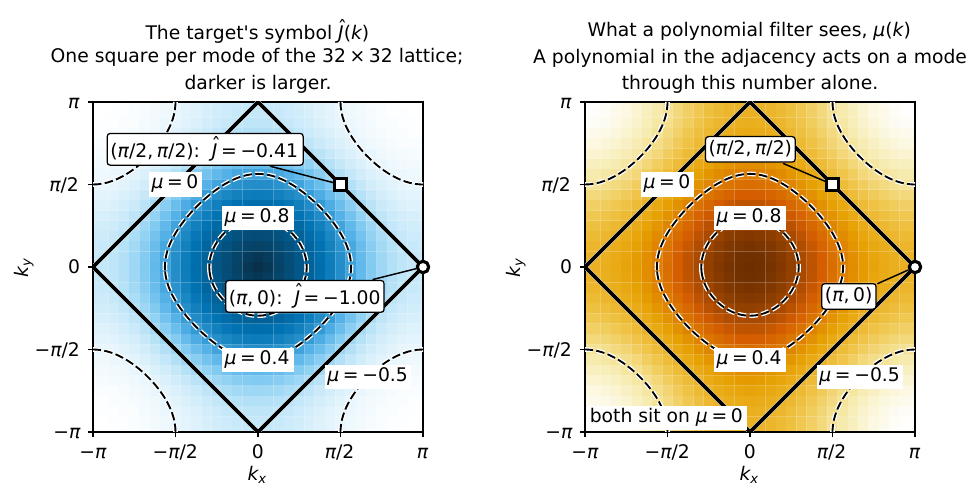}
\caption{Why a polynomial filter leaves a floor. Both panels are the lattice's frequency square. On
the left is the target's symbol, with the level sets of $\mu(k) = (\cos k_x + \cos k_y)/2$ drawn over
it; on the right is $\mu$ itself, the only quantity a filter polynomial in the adjacency can respond
to. Such a filter takes one value along each curve. On this grid of modes the target, too, takes one
value on every level set except $\mu = 0$, where the two marked modes carry target values of \jhatAxisModeLthirtytwo{} and
\jhatDiagonalModeLthirtytwo{}. The weighted variance of the target along that curve is the floor of
\cref{prop:floor}. Drawn at \latticeLthirtytwoNodes{} nodes.}
\label{fig:levelsets}
\end{figure}

A filter that is a polynomial in the lattice adjacency $A$ acts on the Fourier mode $k$ by
multiplication with $p(2\cos k_x + 2\cos k_y)$. Its symbol therefore depends on the frequency only
through the scalar $\mu(k) = (\cos k_x + \cos k_y)/2$, and takes one value on each level set of
$\mu$. The target's symbol $\hat{J}$ does not. The modes $k = (\pi, 0)$ and $k = (\pi/2, \pi/2)$
both have $\mu = 0$, and the coupling gives them different values, because the coupling is
isotropic in space while $\mu$ is not isotropic in frequency. \Cref{fig:levelsets} shows the two
side by side. The best a polynomial filter can do is to take, on each level set, the weighted mean of
the target over that set, and what it cannot remove is the variation of the target within the set.

\begin{proposition}
\label{prop:floor}
Let the kernel $R$ be translation invariant with a symbol that is a function of $\mu$, and let
$S(k) = \mathbb{E}\,\lvert \hat{s}(k)\rvert^2$ be the power spectrum of the spin fields, with the
unnormalized transform $\hat{s}(k) = \sum_x s_x e^{-i k\cdot x}$, the same convention for $\hat{R}$
and $\hat{J}$, and the expectation over configurations. Then the mean squared error of the readout
$\hat{y}_i = s_i \cdot (R \circledast s)_i$ is
\[
\frac{1}{N^{2}}\sum_k S(k)\,\lvert \hat{R}(k) - \hat{J}(k)\rvert^2
\;=\; \sum_m W_m \big(\hat{R}(m) - \bar{J}_m\big)^2
\;+\; \frac{1}{N^{2}}\sum_k S(k)\,\big(\hat{J}(k) - \bar{J}_{\mu(k)}\big)^2 ,
\]
where $\bar{J}_m$ is the $S$-weighted mean of $\hat{J}$ on the level set $\{\mu = m\}$ and
$W_m = N^{-2}\sum_{\mu(k) = m} S(k)$. The second term does not involve $R$. It is a lower bound on
the error of the readout with $R = p(A)$, for every polynomial $p$ of every degree.
\end{proposition}

\begin{proof}
The lattice adjacency is diagonalized by the Fourier basis, with eigenvalue $2\cos k_x + 2\cos k_y$
at mode $k$. A polynomial $p(A)$ therefore acts by multiplication with $p(2\cos k_x + 2\cos k_y)$,
which depends on $k$ only through $\mu(k) = (\cos k_x + \cos k_y)/2$. Because the spins are $\pm 1$
and the target is the convolution in \cref{eq:target}, the mean squared error of the readout
$\hat{y}_i = s_i \cdot (R \circledast s)_i$, with $R$ translation invariant, is
$N^{-2}\sum_k S(k) \lvert \hat{R}(k) - \hat{J}(k)\rvert^2$ with $S$ the power spectrum of the spin
fields. Split the sum over $k$ into the level sets of $\mu$. Inside the level set $\{\mu = m\}$ the symbol
$\hat{R}$ is a constant, so the sum decomposes into the weighted distance from that constant to the
weighted mean $\bar{J}_m$ plus the weighted variance of $\hat{J}$ inside the level set. The first
part vanishes at the choice $\hat{R}(m) = \bar{J}_m$, which a polynomial of degree one less than
the number of distinct values of $\mu$ attains exactly, by interpolation. The second part is free
of $R$.
\end{proof}

The decomposition itself is standard: it writes as a weighted error the observation of
\citet{wang2022powerful} that a polynomial filter acts as one number on each eigenvalue of the
adjacency, which \citet{lu2024eigenvalue} tie to the number of distinct eigenvalues. What is new here is that on these lattices the floor can be
computed exactly, because \cref{lem:onelevel} below reduces it to the variation of the target along a
single level set. The gate by $s_i$ is part of the class: a linear filter applied to $s$ without it
cannot represent the target at all. The proof also says how to compute the floor:
form the weighted variance of $\hat{J}$ within the level sets of $\mu$, using the spin spectrum of the
training split as weights. The floor depends on the data through $S$, so it is a property
of the configurations as well as of the lattice. We compute it with the spin spectrum of
\polySpectrumGraphs{} training configurations, for which it is exact, and the spectrum of the test
configurations moves it by at most \polyFloorTestShiftMax{}. It is
\polyFloorLthirtytwoRound{} at \latticeLthirtytwoNodes{} nodes and lies between
\polyFloorShallowRound{} and \polyFloorDeepRound{} across the five sizes (\cref{tab:closed}). All of it comes from one
level set, because of a number-theoretic property of the lattice when the side $L$ is a power of two,
as at every size of the benchmark.

\begin{table}[tbp]
\centering
\scriptsize
\setlength{\tabcolsep}{4pt}
\providecommand{\RrefOneLevel}{\cref{app:proofs}}%
\textbf{(a) How far the filter has to reach} (least squares against the known kernel, weighted by the spin spectrum of the official training split, $L=32$).

\begin{tabular}{ccc}
\toprule
reach radius & coefficients & logMSE \\
\midrule
$4$ & 9 & ${-}0.665$ \\
$8$ & 31 & ${-}1.688$ \\
$16$ & 113 & ${-}3.659$ \\
$22.7$ & 152 & ${-}27.410$ \\
\bottomrule
\end{tabular}

\vspace{1em}
\textbf{(b) Polynomial filters in the adjacency}: the best fit at each degree, the floor, and how many distinct values the symbol takes.

\begin{tabular*}{0.9\linewidth}{@{\extracolsep{\fill}}lccccccc}
\toprule
Size & degree 8 & 16 & 32 & 64 & 128 & degree $L$ & floor (any degree) \\
\midrule
$L=16$ & ${-}1.636$ & ${-}1.973$ & ${-}2.306$ & ${-}2.475$ & ${-}2.475$ & ${-}1.973$ & \textbf{${-}2.475$} \\
$L=32$ & ${-}1.179$ & ${-}1.817$ & ${-}1.921$ & ${-}2.050$ & ${-}2.458$ & ${-}1.921$ & \textbf{${-}2.678$} \\
$L=64$ & ${-}1.012$ & ${-}1.619$ & ${-}1.884$ & ${-}1.907$ & ${-}1.977$ & ${-}1.907$ & \textbf{${-}2.949$} \\
$L=128$ & ${-}0.934$ & ${-}1.509$ & ${-}1.827$ & ${-}1.893$ & ${-}1.898$ & ${-}1.898$ & \textbf{${-}3.233$} \\
$L=256$ & ${-}1.022$ & ${-}1.498$ & ${-}1.787$ & ${-}1.872$ & ${-}1.885$ & ${-}1.886$ & \textbf{${-}3.518$} \\
\midrule
\multicolumn{8}{@{}l@{}}{\rlap{\parbox[t]{0.9\linewidth}{\footnotesize Degree $L$ is the diameter of the lattice, the degree at which the filter first reaches every node. Distinct symbol values: 41 at $L=16$, 145 at $L=32$, 545 at $L=64$, 2\,113 at $L=128$, 8\,321 at $L=256$. At $L=16$ every degree from \polyExactDegreeLsixteen{} on attains the floor, so the degree-64 and degree-128 cells equal it.}}} \\
\bottomrule
\end{tabular*}

\vspace{1em}
\textbf{(c) Leaving the strided basis}: what each change to the basis is worth, by least squares against the known kernel.

\begin{tabular*}{0.9\linewidth}{@{\extracolsep{\fill}}lcccc}
\toprule
Basis & \multicolumn{2}{c}{$L=32$} & \multicolumn{2}{c}{$L=64$} \\
\cmidrule(lr){2-3}\cmidrule(lr){4-5}
 & coefficients & logMSE & coefficients & logMSE \\
\midrule
the basis itself, five-tap filter & 49 & ${-}6.567$ & 55 & ${-}6.972$ \\
stencil radius 3 & 57 & ${-}6.699$ & 67 & ${-}7.514$ \\
stencil radius 4 & 67 & ${-}6.723$ & 82 & ${-}7.594$ \\
fine reach 12 & 83 & ${-}6.755$ & 89 & ${-}7.407$ \\
fine reach 16 & 131 & ${-}7.666$ & 137 & ${-}7.466$ \\
three-tap filter & 49 & ${-}5.108$ & 55 & ${-}5.035$ \\
seven-tap filter & 49 & ${-}6.606$ & 55 & ${-}7.427$ \\
denser ladder & 49 & ${-}6.546$ & 55 & ${-}7.348$ \\
radius 3 + fine 12 & 91 & ${-}6.901$ & 101 & ${-}7.609$ \\
\midrule
\multicolumn{5}{@{}l@{}}{\rlap{\parbox[t]{0.9\linewidth}{\footnotesize Filter width is the largest lever: going from five taps to three costs $1.94$ in logMSE at $L=64$ and $1.46$ at $L=32$. The variants that widen the basis land between ${-}7.35$ and ${-}7.61$ at $L=64$ and between ${-}6.55$ and ${-}7.67$ at $L=32$; the three-tap row is above both bands.}}} \\
\bottomrule
\end{tabular*}

\vspace{1em}
\textbf{(d) Where the floor comes from}: the level sets of $\mu$, one row per size, from the same weighted decomposition that gives the floor.

\begin{tabular*}{0.9\linewidth}{@{\extracolsep{\fill}}lccccc}
\toprule
Size & distinct values of $\mu$ & modes on $\mu=0$ & share of all modes & share of the floor & floor \\
\midrule
$L=16$ & 41 & 30 & $11.72\%$ & $100\%$ & ${-}2.475$ \\
$L=32$ & 145 & 62 & $6.05\%$ & $100\%$ & ${-}2.678$ \\
$L=64$ & 545 & 126 & $3.08\%$ & $100\%$ & ${-}2.949$ \\
$L=128$ & 2\,113 & 254 & $1.55\%$ & $100\%$ & ${-}3.233$ \\
$L=256$ & 8\,321 & 510 & $0.78\%$ & $100\%$ & ${-}3.518$ \\
\midrule
\multicolumn{6}{@{}l@{}}{\rlap{\parbox[t]{0.9\linewidth}{\footnotesize Because $L$ is a power of two, the target is constant along every level set of $\mu$ except the one through $\mu=0$ (\RrefOneLevel), which holds $2L-2$ of the $L^2$ modes, so the whole floor is its variation along that one curve. The largest term from any other level is below $4\times 10^{-29}$ of the floor, and within those levels the target varies by at most $4\times 10^{-15}$, both at the size of the arithmetic's own error. Doubling the lattice moves the floor by ${-}0.20$, ${-}0.27$, ${-}0.28$, ${-}0.28$, against the ${-}0.29$, ${-}0.29$, ${-}0.30$, ${-}0.30$ that the shrinking share of the modes alone would give.}}} \\
\bottomrule
\end{tabular*}

\vspace{1em}
\textbf{(e) The strides one level at a time}: the kernels of part (c)'s first row built up from its fine offsets, one level of strides at a time, the last with its farthest taps half the lattice away; least squares against the known kernel, weighted by the spin spectrum of the official training split, in double precision. The full basis has rank 43 and 48 (\cref{sec:strides}), and lands within $0.01$ of part (c)'s first row, which used a different solver. Gains are in orders of magnitude.

\begin{tabular*}{0.9\linewidth}{@{\extracolsep{\fill}}lcccccc}
\toprule
 & \multicolumn{3}{c}{$L=32$} & \multicolumn{3}{c}{$L=64$} \\
\cmidrule(lr){2-4}\cmidrule(lr){5-7}
Farthest tap & coefficients & logMSE & gain & coefficients & logMSE & gain \\
\midrule
fine offsets only & 31 & ${-}1.681$ & -- & 31 & ${-}1.392$ & -- \\
$4$ & 37 & ${-}1.700$ & $0.02$ & 37 & ${-}1.406$ & $0.01$ \\
$8$ & 43 & ${-}2.918$ & $1.22$ & 43 & ${-}2.146$ & $0.74$ \\
$16$ & 49 & ${-}6.567$ & $\mathbf{3.65}$ & 49 & ${-}3.655$ & $1.51$ \\
$32$ & -- & -- & -- & 55 & ${-}6.972$ & $\mathbf{3.32}$ \\
\midrule
\multicolumn{7}{@{}l@{}}{\rlap{\parbox[t]{0.9\linewidth}{\footnotesize In bold, the last level, which adds directions the finer levels do not span. Gains depend on the order in which levels are added; \cref{app:closedforms} gives the cost of leaving each one out. The gain of that level is $3.65$ at $L=32$, $3.32$ at $L=64$; the shorter levels together give $1.24$ at $L=32$, $2.26$ at $L=64$.}}} \\
\bottomrule
\end{tabular*}

\caption{The closed forms behind \cref{sec:benchmark} and \cref{sec:floor}: how far the filter has
to reach, the degree ladder for polynomials in the adjacency with its floor, what each change to the
strided basis is worth, where the floor comes from, and what each level of strides adds. Every part
uses the known coupling, not the labels, and the spin spectrum of \polySpectrumGraphs{} training
configurations. The text describes every part.}
\label{tab:closed}
\end{table}

\begin{lemma}
\label{lem:onelevel}
Let $L \ge 4$ be a power of two. If two modes of the $L \times L$ torus share a nonzero value of
$\mu$, one of the eight symmetries of the square maps one to the other.
\end{lemma}

\begin{proof}
Let $\zeta = e^{2\pi i/L}$. The
powers $1, \zeta, \ldots, \zeta^{L/2-1}$ are a basis of $\mathbb{Q}(\zeta)$ over the rationals, and
$2\cos(2\pi j/L) = \zeta^{j} - \zeta^{L/2-j}$ for $0 < j < L/2$. Every $\cos(2\pi j/L)$ is
therefore $0$ or $\pm c_i$ for one of $c_0 = 1$ and $c_i = \cos(2\pi i/L)$ with $0 < i < L/4$, and
the $c_i$ are linearly independent over the rationals because they use disjoint elements of the
basis. A sum of two such numbers that is not zero determines its two terms. So two modes with the
same nonzero $\mu$ have the same pair of cosines, which on the grid means that one is the image of
the other under one of the eight symmetries of the square.
\end{proof}

The target has those symmetries, since the coupling depends only on the torus distance, so it is constant
along every level set except the one through $\mu = 0$, and the floor is exactly its variation along
that single curve. In the
terms of \citet{wang2022powerful}, zero is the only eigenvalue of the adjacency whose multiplicity
the symmetries do not explain. At some sides that are not powers of two other sums coincide, for
instance $\cos 0 + \cos(2\pi \cdot 4/12) = \cos(2\pi \cdot 2/12) + \cos(2\pi \cdot 3/12)$ at
$L=12$, and such level sets add to the floor. The curve holds $2L-2$ of the $L^2$ modes, and the floor
falls about as $1/L$, although the weights on the curve and the target along it also change with the size. It falls by \polyFloorStepLow{} to \polyFloorStepHigh{} in logMSE per doubling of
the side, against \floorStepInverseL{} for a law of $1/L$.

\begin{figure}[t]
\centering
\includegraphics[width=\linewidth]{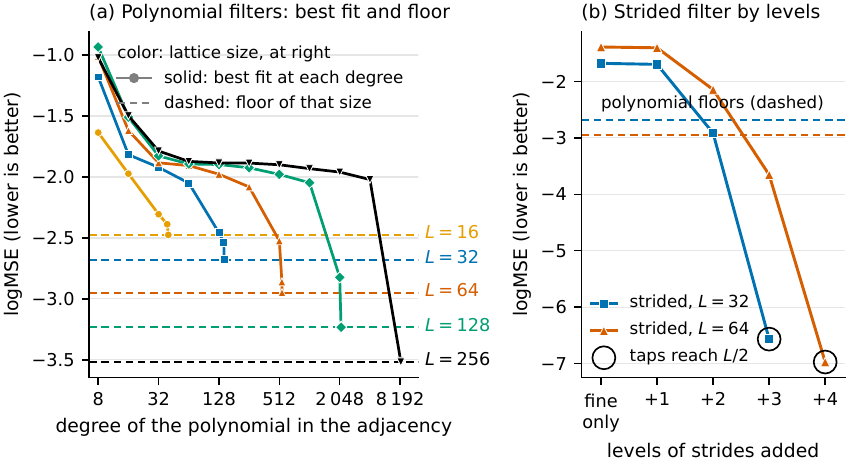}
\caption{Fits to the known coupling, weighted by the spin spectrum of \polySpectrumGraphs{} training
configurations. (a) The best filter that is a polynomial in the adjacency, against its degree, one line
per lattice size; the dashed line in the same color is that size's floor (\cref{prop:floor}). At every
size the fit reaches the floor at the last degree, one less than the number of distinct values of
$\mu$, which runs from \polyLevelsLsixteen{} to \polyLevelsLtwofiftysix{}, and at the three larger
sizes it first levels off above the floor. Markers are the degrees at which the fit was computed,
and the lines join them; the best fit cannot rise as the degree grows. (b) The strided filter of
\cref{sec:strides} fitted to the coupling, starting from its fine offsets and adding one level of strides
at a time (\cref{tab:closed}, part e), at the two sizes where that ladder was computed, with the
polynomial floor at each size dashed in the same color. The circled points are the full bases, whose top
level reaches half the lattice; the gain of a level depends on the order in which levels are added.
Lower is better in both panels.}
\label{fig:closedforms}
\end{figure}

Reaching the floor is expensive. Interpolation attains it at a degree one less than the number of
distinct values of $\mu$, which runs from \polyLevelsLsixteen{} to \polyLevelsLtwofiftysix{}, about an eighth of the
number of nodes. A polynomial of degree \polyExactDegreeLsixteen{} attains it at $L=16$, and at the
larger sizes the fit comes within \polyWithinTol{} of the floor only close to that degree: at degree
\polyWithinDegreeLsixtyfour{} at $L=64$, where interpolation needs \polyExactDegreeLsixtyfour{}, and
at \polyWithinDegreeLonetwentyeight{} at $L=128$, where it needs \polyExactDegreeLonetwentyeight{}. A
polynomial whose degree equals the diameter of the lattice, which already reaches every node, is
near \polyDiameterRound{} at every size (\cref{tab:closed}, part b, and \cref{fig:closedforms}a), and
the fit falls to the floor only as the degree grows toward the interpolation degree. These fits use a
basis that stays exact at every degree. In a monomial basis the design matrix loses rank as the degree
grows and the fit stops improving near \priorPolySaturation{}, so that this value looks like the limit
(\cref{app:proofs}).

The floor binds one class: the linear propagation of polynomial spectral networks such as ChebNet
\citep{defferrard2016chebnet}, GPR-GNN \citep{chien2021gprgnn} and BernNet \citep{he2021bernnet},
used inside the target's gate, that is, the readouts $s_i\,(p(A)\,s)_i$ whose filter is a polynomial
in the adjacency, and spectral networks of that class adapted from published ones, fitted to the benchmark, land where
the closed forms say (\cref{sec:spectral}). A network with a nonlinearity after each round is not bound by it, so the floor does
not predict its score, and the measured networks fall on both sides of it. An eight-layer graph
network with mean aggregation, which we call the eight-layer GCN and trained end to end at the three
larger sizes with the step budget of \cref{sec:setup}, scores within \gapBaselineToDegreeEightMax{} of
the best polynomial of degree eight (\cref{tab:closed}, part b), and \gapBaselineToCapLowRound{} to
\gapBaselineToCapHighRound{} orders of magnitude worse than the floor. That network has no gate and a
nonlinearity in every layer, so we record the agreement with degree eight as an observation; we did
not test whether it is more than a coincidence. Our SIGN readout with eight rounds of propagation, which is also nonlinear, lands \specSignEightBelowLow{} to \specSignEightBelowHigh{} below the best polynomial of that degree
(\cref{sec:spectral}), as we had expected before training it. Among the published networks, a plain
message-passing network \citep{xu2019gin} sits above the floor at \latticeLthirtytwoNodes{} nodes, at
\pubGinLthirtytwo{}, while a gated one \citep{bresson2017gated}, at \pubGatedGcnLthirtytwo{}, and the
attention models \citep{rampasek2022gps}, at \pubGpsLthirtytwoHigh{} to \pubGpsLthirtytwoLow{}, and the hierarchical
recurrent network of \citet{carotti2026ghr}, at \pubGhrLthirtytwo{}, score below it, so the floor does not account for where those networks stop.

The floor binds filters that are functions of $\mu$, and a filter that is not one is not bound by it. The
adjacency ties offsets together. $A^2$ reaches the offset $(2,0)$, two steps along an axis, and the
offset $(1,1)$, one step along each axis, in the fixed ratio of one to two; $A^4$ weighs the same two
offsets sixteen to twenty-four, so combinations of powers can change that ratio, but only together
with farther offsets, and every combination still takes one value on each level set of $\mu$. That
is the precise sense in which the adjacency lumps offsets together. A free $3\times3$ stencil is not
bound in this way: its diagonal part, the mean of the four shifts by $(\pm1,\pm1)$, has the symbol
$\cos k_x \cos k_y$, which is $-1$ at $(\pi,0)$ and $0$ at $(\pi/2,\pi/2)$, two modes on the level
set $\mu = 0$. A shift by $d$ sites along an axis gives a new filter for each $d$: the mean of the shifts by $\pm d$ over the two
axes has the symbol $(\cos d k_x + \cos d k_y)/2$, which is a different function of $k$ for every
$d$ from $1$ to $L/2$, since the shifts by $d$ and $L-d$ coincide. On these lattices it is a function of $\mu$ for odd $d$, since it vanishes on the level set
$\mu = 0$ and the other level sets are orbits of the symmetries, and it is not one for even $d$.

\subsection{How far the filter must reach, and what strides add}
\label{sec:strides}

How far must a filter that is free of the floor reach, and how can it get there cheaply? Stacks of
$3\times3$ convolutions span every
translation-invariant operator on the torus, so as a class they have no floor at all, but they need
reach: only a kernel that covers the torus reproduces the coupling (\cref{sec:benchmark} and
\cref{tab:closed}, part a), and a
stack of $3\times3$ convolutions covers the torus only after about $L/2$ layers.

\begin{figure}[t]
\centering
\includegraphics[width=\linewidth]{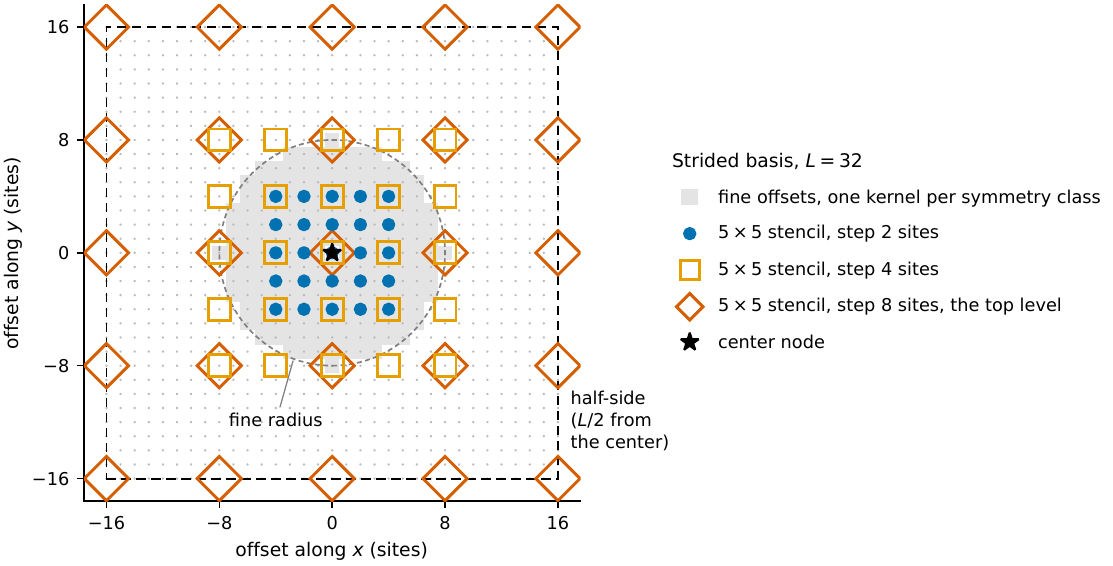}
\caption{The fifty-coefficient strided basis on the lattice of \latticeLthirtytwoNodes{} nodes, drawn around one center node. The
shaded sites inside the dashed circle are the fine offsets, which the basis groups into one kernel per
class of offsets related by a symmetry of the square. Each set of markers is one level of strides: a
five-by-five stencil whose step doubles from level to level, so that the farthest taps of the top level
lie on the dashed square at half the side of the lattice. On the torus the opposite edges of that square
are the same sites, and the taps drawn on both edges are one set. Where levels share a site, the markers
are nested. The markers are the centers of the taps only: each level's kernels are smoothed by a
binomial filter that widens with the level, so the coarse levels spread their weight over a wide patch
around each marked site. With the fine offsets, these levels give \atrousCoefThirtyTwo{} kernels, of
rank \atrousRankThirtyTwo{}, which a least-squares fit weighs separately.}
\label{fig:stridedbasis}
\end{figure}

Strides give that reach in a number of levels that grows with $\log L$. Our strided basis
(\cref{fig:stridedbasis}) holds the
symmetric offsets within radius eight and, at each level of a ladder whose step doubles from two
sites to a quarter of the lattice, the symmetry classes of a five-by-five stencil with that step,
smoothed, so that the farthest taps of the top level lie half the lattice away. In full, every kernel
of the basis is taken on the torus and is invariant under
the eight symmetries of the square. Let $\rho$ be the fine radius, eight for the fifty-coefficient
filter and twelve for the larger one. The fine part holds one kernel for each orbit of
the symmetries on the offsets $a$ with $0 < \lVert a \rVert_2 \le \rho$, equal to one on every offset of
the orbit and zero elsewhere. Let $h = (1, 4, 6, 4, 1)/16$, and let $h_j$ be the separable
two-dimensional filter $h \otimes h$ with its taps $2^j$ sites apart. Level $l$, for
$l = 0, \ldots, \log_2 L - 3$, holds six kernels: the indicators of the six orbits of the $5\times5$
stencil $\{-2, \ldots, 2\}^2$, its center included, with the stencil's step stretched to $2^{l+1}$
sites, each convolved with $h_0 * h_1 * \cdots * h_l$. The steps therefore double from two sites to
$L/4$, and the farthest taps of level $l$ lie $2^{l+2}$ sites from the origin along an axis, out to
$L/2$ at the top level. A kernel that is zero, or a multiple of an earlier one, is dropped; no kernel is
normalized, and a least-squares fit does not depend on the scale of a kernel. This gives
\atrousCoefThirtyTwo{} and \atrousCoefSixtyFour{} kernels at $L=32$ and $L=64$ for the
fifty-coefficient filter, and \largeStridedCoefLow{} to \largeStridedCoefHigh{} from $L=32$ to $L=256$ for
the larger one. Fewer of them are independent. The center kernel of level $l$ is the
smoothing $h_0 * \cdots * h_l$ itself, and since $h_l$ is a combination of the orbit indicators of
the stencil at level $l-1$'s step, that kernel is a combination of level $l-1$'s six kernels. Each
level therefore adds at most five directions, and the first adds fewer, since it overlaps the fine
offsets. The fifty-coefficient basis has rank \atrousRankThirtyTwo{} at $L=32$ and
\atrousRankSixtyFour{} at $L=64$, and the larger basis spans
\readoutIndepDirsLthirtytwo{} to \readoutIndepDirsLtwofiftysix{} directions from $L=32$ to $L=256$.
The steps stop at a quarter of the lattice because at half of it the shifts in the two directions
along an axis coincide.

\begin{remark}[The half-lattice stride]
\label{app:halfstride}
On a cycle of length $L$ with $L$ even, the shift by $+L/2$ and the shift by $-L/2$ are the same
permutation, because $L/2 \equiv -L/2 \pmod{L}$. On the torus the same holds along each axis. A map
that assigns a separate operator to each of the four directions therefore applies two of its
operators, say $W_{+}$ and $W_{-}$, to the same shifted field $R_{L/2}\,z$ when the stride is $L/2$,
and the pair contributes $(W_{+} + W_{-})\,R_{L/2}\,z$. Any term at that stride that one operator
can produce, the isotropic term included, is therefore within reach of such a map; what it loses
there is the separation of the two directions. The coincidence is exactly $2d \equiv 0 \pmod{L}$, so
strides other than $L/2$ stay distinct even when they exceed $L/4$; what stops a ladder whose strides
double at a quarter of the lattice, if its maps must stay distinct, is that its next step would be
$L/2$.
\end{remark}

Used inside the target's gate, the basis gives a readout of the target's form, and we fit its
coefficients in two ways. Fitted to the coupling, the coefficients $c_p$ of the basis kernels $B_p$
minimize $N^{-2}\sum_k S(k)\,\lvert \sum_p c_p \hat{B}_p(k) - \hat{J}(k)\rvert^2$, which by the
identity of \cref{prop:floor} is the mean squared error of the readout with kernel $\sum_p c_p B_p$,
with $S$ the spin spectrum of \polySpectrumGraphs{} training configurations. We solve it in double
precision as ordinary least squares on the design matrix weighted by $\sqrt{S}$, with a solver based
on the singular value decomposition, and report the residual itself. These values are exact for the
training spectrum and are not scores on the test split. Fitted to the training labels, the
coefficients minimize $\sum_i \big(y_i - s_i \sum_p c_p (B_p \circledast s)_i\big)^2$ over every node
of the training split, with no constant. We accumulate a QR factorization over the training graphs one
at a time, solve every sub-basis (fine radius two to twelve in steps of two, any number of levels) from the same
factor, choose the sub-basis on the validation split, and read the test split once. Validation
chooses fine radius twelve with every level at each size, which is the larger filter.
The fifty-coefficient filter fitted to the labels is the sub-basis of fine radius eight with every
level, solved from the same factor and read once on the test split.

The basis's smoothing width, stencil size and ladder were chosen on fits to the known coupling
(\cref{tab:closed}, part c); only its coefficients come from the labels. In part (c), the three-tap
and seven-tap rows replace the five-tap smoothing filter $h$, a stencil radius of $r$ uses the stencil
$\{-r, \ldots, r\}^2$ in place of the $5\times5$ one, the denser ladder stretches the stencil's step to
$2^l$ instead of $2^{l+1}$ sites at level $l$, and the fine-reach rows set $\rho$. Fitted to the training
labels, its fifty or so coefficients reach \atrousLabelTestLthirtytwoRound{} on the test split at
\latticeLthirtytwoNodes{} nodes (\cref{tab:compact}) and
\atrousLabelTestLsixtyfourRound{} at \latticeLsixtyfourNodes{} nodes, where no published model was
trained. Fitted to the coupling, the top level of strides is worth the most: left out of the full basis it
costs \atrousDropTopCostLthirtytwo{} and \atrousDropTopCostLsixtyfour{} in logMSE at the two sizes,
more than any other level (\cref{app:closedforms}). Added one at a time from the finest, the last
two levels give most of the gain (\cref{fig:closedforms}b). With its step halved, so that its farthest taps
stop at a quarter of the lattice, the basis does as well, because the smoothing still spreads each
kernel across the lattice (\cref{tab:stand}); what the coupling asks of such a filter is resolution
out to the whole lattice. With offsets out to radius twelve (\largeStridedCoefLow{} to
\largeStridedCoefHigh{} coefficients), the same fit to the labels does better still; this larger strided
filter gives the row ``The larger strided filter'' of \cref{tab:compact}.

For a filter of the target's form on this lattice, then, there are two requirements. It must weigh
separately the offsets the adjacency lumps together, which no degree of a polynomial filter does, and it
must resolve the coupling from the nearest neighbors out to the whole lattice, which a strided basis
does with a few dozen coefficients.

\section{Experimental setup}
\label{sec:setup}

How were the networks trained and scored, and how does that differ from the benchmark's published
protocol? For each experiment we wrote down, before it ran, what we expected and what would count
against the expectation; \cref{app:registered} gives each expectation and what happened.

\subsection{Data and protocol}
\label{sec:data}

\paragraph{The published protocol.} The benchmark holds \benchConfigs{} configurations per dataset
under an \benchSplit{} split, at the two coupling exponents \benchSigmaHard{} and \benchSigmaEasy{},
and reports the base-ten logarithm of the mean squared error. It trains for up to \benchEpochs{}
epochs with a patience of \benchPatience{} epochs, caps each model at \benchHourCap{} hours of
compute and reports \benchSeedsWord{} models per number. The benchmark paper reports no parameter
counts or peak memory.

\paragraph{Our data.} Every number we report for a trained model uses the official files and the
official \benchSplit{} split, at the harder exponent except in \cref{sec:secondcoupling}. The split
takes the first \nTrainGraphs{} configurations for training, the next \nTestGraphs{} for validation and the last \nTestGraphs{} for test. Every
network of \cref{sec:archs}, the eight-layer GCN included, trains on all \nTrainGraphs{}
training configurations at every size, the largest included, and every test number is measured on
all \nTestGraphs{} test graphs. Three computations use fewer training graphs, and none of them
trains one of the standard architectures: the learned readout at $L=256$, described in \cref{sec:archs}; the spin spectrum behind the floor
of \cref{prop:floor} and the polynomial fits of \cref{tab:closed}, computed from
\polySpectrumGraphs{} training configurations at each size; and the check of the operator identity
(\cref{app:proofs}).

\paragraph{Our protocol.} In place of the benchmark's epochs and patience, every network trains,
unless stated otherwise, with Adam at \baseLrRegistered{}, a cosine schedule and one step budget per size. For the standard architectures no checkpoint or
hyperparameter is selected: each model is trained once and scored on all \nTestGraphs{} official test
graphs, and each entry is the mean and sample standard deviation of three models. The polynomial
spectral networks of \cref{sec:archs} have a budget of their own, given with them.

\subsection{Architectures}
\label{sec:archs}
\label{app:baselines}

\paragraph{The dilated networks.} The strided model is a dilated convolutional network \citep{yu2016dilated} in the residual form of
\citet{yu2017drn}. Each residual block holds two $3\times3$ circular convolutions at one
dilation. The dilations double from one site to a quarter of the lattice, because at half the lattice
the shifts in the two directions along an axis coincide (\cref{app:halfstride} in \cref{sec:strides}), and the ladder is
traversed twice, so the network has \dcnnBlocksLow{} to \dcnnBlocksHigh{} blocks from the smallest size to the
largest and a receptive field that covers the torus at every size. Its width is set at each size to
give about two million parameters, and a small version keeps the depth at \dcnnSmallFractionWord{} of
the parameters. In full, the network is a $1\times1$ lift to width $c$; one pre-norm residual block per
dilation, each a per-site layer normalization over channels, a $3\times3$ convolution, a GELU and a
second $3\times3$ convolution, both convolutions with circular padding and the block's dilation; and
a per-site layer normalization and a two-layer $1\times1$ head. \Cref{tab:hyper} gives the width
and the parameter count of both versions at each size. The gated version of
\cref{sec:distance} keeps the large network unchanged at $L=16$ and $L=32$ and reads its head's output
$c_i$ as $b + s_i c_i$, with one learned scalar $b$; it was trained once at each size with the same recipe.

\paragraph{The Fourier neural operators.} The model with global reach is a Fourier
neural operator \citep{li2021fno} in the configuration its authors publish for two-dimensional
problems: a pointwise lift to width \fnoWidth{}; \fnoLayersWord{} Fourier layers, each the sum
of a spectral convolution on the lowest \fnoModes{} modes per axis and a pointwise linear path, with a
GELU after all but the last; and a two-layer pointwise projection, with \fnoParamsLowM{} million
parameters. Three choices follow the authors'
maintained library rather than the paper's text, and each is stated here. We use GELU without batch
normalization, as the library does. We add no domain padding, since the lattice is periodic. We
append no grid coordinates, since the target is exactly translation invariant on the torus and
coordinates that are not periodic can only break that symmetry. At $L=16$ the lattice holds fewer than \fnoModes{}
modes per axis, so the operator keeps all of them, and its width is \fnoWidthLsixteen{} to keep its
parameter count. As for the dilated network, a small version has about \fnoSmallFractionWord{} of the
parameters, at width \fnoSmallWidth{}. To separate truncation from the architecture we also trained,
at \latticeLthirtytwoNodes{} and \latticeLsixtyfourNodes{} nodes, an operator that keeps every mode
at about the same parameter count, with width \fnoAllModesWidth{} and \fnoAllModesParamsM{} million
parameters at the first size, and at \latticeLthirtytwoNodes{} nodes one with the two grid channels
of the reference code, which tests the choice of no coordinates. All of these networks, the dilated ones included, map the spin field, read as an
image, to the per-node label, and none carries the target's quadratic form or the coupling.

\paragraph{The gated graph networks.} Each layer is the gated graph convolution of
\citet{bresson2017gated} as GraphGPS implements it \citep{rampasek2022gps}, with residual connections.
For the edge from $j$ to $i$, with node state $x$ and edge state $e$, it forms
\[
\hat{e}_{ij} = D x_i + E x_j + C e_{ij}, \qquad
\hat{x}_i = U x_i + \frac{\sum_j \sigma(\hat{e}_{ij}) \odot V x_j}{\sum_j \sigma(\hat{e}_{ij}) + \epsilon},
\]
with $\sigma$ the logistic function and $\epsilon$ a small constant, and updates
$x_i \leftarrow x_i + \mathrm{drop}(\mathrm{ReLU}(\mathrm{BN}(\hat{x}_i)))$ and
$e_{ij} \leftarrow e_{ij} + \mathrm{drop}(\mathrm{ReLU}(\mathrm{BN}(\hat{e}_{ij})))$; every edge state
is one before the first layer. As in GraphGPS, a graph-level layer normalization and a two-layer
feed-forward block with its own residual connection and normalization follow, and the head is a linear
layer, a ReLU, an $\ell_2$ normalization and a linear readout. In the per-direction network $V$ and
$E$ have a separate matrix for each of the four edge directions, and $U$, $C$ and $D$ are shared. The
strided network adds, in layer $\ell$, four more edge types at the stride $s_\ell$ along each axis,
each with its own $V$ and $E$, where $s_\ell$ cycles through $2, 4, \ldots, L/2$ from layer to layer;
at stride $L/2$ the two opposite edges join the same pair of nodes and stay two types. The stride-one
edges carry their state from layer to layer, and the strided edges of each layer start from ones.
The attention model adds, in each layer, multi-head attention over all nodes, as the benchmark
configures GraphGPS with random-walk encodings. Each network runs at the depth of its benchmark
configuration, the per-direction and strided ones at the GatedGCN's, with the width set for about two
million parameters (\cref{tab:graphmodels}).

\paragraph{The eight-layer GCN.} We call it a GCN for brevity, but it is a message-passing network
with mean aggregation and a separate weight for the node itself, not the normalized convolution of
\citet{kipf2017gcn}. A linear map and a SiLU lift the spin to width \gcnEightWidth{}, eight layers
update $h_i \leftarrow \mathrm{LayerNorm}\big(h_i + \mathrm{SiLU}(W_\ell [h_i, \bar{h}_i] + b_\ell)\big)$,
where $\bar{h}_i$ is the mean of $h$ over the four neighbors of $i$, and a linear map reads out the
label. It has \gcnEightParamsM{} million parameters, and its linear part after eight layers is a polynomial of
degree eight in the adjacency.
It was trained end to end on the official split at $L=64$, 128 and 256, two models per size, and
evaluated on the official test split. It ran the same steps and batches as the other baselines at
those sizes, \gcnEightSteps{} steps at $L=64$ and $L=128$ and \gcnEightStepsLtwofiftysix{} at a batch
of \gcnEightBatchLtwofiftysix{} at $L=256$, with Adam at a constant rate of \gcnEightLr{} and no
clipping. Its peak training memory is about \gcnEightMemRoundGB{} GB at $L=64$ and $L=128$ and
\gcnEightMemLtwofiftysixRoundGB{} GB at $L=256$.

\paragraph{The polynomial spectral networks.} Two spectral networks adapted from published architectures, built on the
linear part that \cref{prop:floor} addresses, were fitted or trained at \benchNodesLow{}, \latticeLthirtytwoNodes{} and
\latticeLsixtyfourNodes{} nodes, on CPU. Both read the channels $z_k = T_k(A/4)\,s$ for
$k = 0, \ldots, K$, where $T_k$ is the Chebyshev polynomial of degree $k$. The channel $z_k$ has the
symbol $T_k(\mu)$, and together the channels span the same space as $s, As, \ldots, A^K s$ while
staying well conditioned at the degrees used here. The gated linear filter
$\hat{y}_i = s_i \sum_{k} \theta_k z_{k,i}$ is the propagation of GPR-GNN \citep{chien2021gprgnn}
inside the target's gate, with Chebyshev channels in place of powers of the normalized adjacency and no network
before the propagation; GPR-GNN itself trains its weights together with such a network. Its constant is absorbed in $\theta_0$, since $s_i z_{0,i} = 1$, and it is
fitted by least squares on the \nTrainGraphs{} training configurations, so it has no seeds. Our SIGN readout follows \citet{frasca2020sign} in reading precomputed
channels, but where SIGN transforms each channel separately before mixing them, ours is a single multilayer perceptron applied at each node to
$[s_i, z_{0,i}, \ldots, z_{K,i}]$, in which $z_{0,i} = s_i$ repeats the spin, with two hidden layers of width \specSignWidth{} and ReLU, and
\specSignParamsLow{} to \specSignParamsHigh{} parameters. It sees $s_i$, so it can learn the gate and
any nonlinear function of the channels. Each SIGN model trains for \specSignSteps{} Adam steps on
batches of \specSignBatch{} nodes from the training split, at a learning rate of \specSignLr{} with a
linear warmup over the first \specSignWarmup{} steps and a cosine decay to zero. Its inputs are
standardized with the statistics of the first \specSignInputGraphs{} training configurations and its
labels with those of the whole training split. We fixed this budget before any run on the official
data, trained three models with different seeds for each setting, and read the last checkpoint once
on the test split. Both networks run at $K=8$, the depth of the eight-layer GCN, and at $K=L$, the
diameter of the lattice.

\paragraph{The quadratic readout.} The readout of \cref{eq:readout} was trained with two choices of
the filter $K$: a free Fourier multiplier, and the basis of the larger strided filter
(\cref{sec:strides}), trained here by gradient descent instead of least squares. Both used Adam with a cosine schedule at a learning rate of
\readoutLR{}, targets standardized and results reported in the benchmark's own units,
\readoutStepsSmall{} steps for the strided filter and \readoutStepsFree{} for the free multiplier
(\readoutStepsFreeLtwofiftysix{} at $L=256$), on \readoutCoresWord{} CPU cores. At $L=256$ the official
training file did not fit in the memory of the CPU machine that trained the readout, so it was
trained on the last \readoutTrainSubsetLtwofiftysix{} configurations of the official training split;
selection and the reported number use the full official validation and test splits at every size.
The free multiplier has \readoutFreeLsixteenParams{} to \readoutFreeLtwofiftysixParams{} parameters and
the strided filter \readoutAtrousLsixteenParams{} to \readoutAtrousLtwofiftysixParams{}. Trained by
gradient descent, the strided filter comes within \readoutAtrousToLSMax{} of the same basis fitted
to the training labels by least squares at every size from $L=32$ on, always on the worse side. At $L=16$ the same basis spans
\readoutIndepDirsLsixteen{} independent directions, which is the dimension of the whole symmetric
kernel space on a 16-torus, so there it is not a strided filter at all but a free symmetric kernel,
and \cref{tab:compact} leaves the strided cell empty at that size. At the other sizes the
independent directions number \readoutIndepDirsLthirtytwo{} to \readoutIndepDirsLtwofiftysix{}, out of
symmetric kernel spaces of \symKernelDimLthirtytwo{} to \symKernelDimLtwofiftysix{} dimensions.

\subsection{Training and scoring}
\label{sec:training}

Labels are standardized with the mean and standard deviation of the training split, batches are drawn
with replacement from the training split, and the loss is the mean squared error on standardized
labels, in single precision. \Cref{tab:hyper} gives the step budget and batch at each size; the
cosine schedule runs to zero, and the gradient norm is clipped at \baseClip{}. Counted in passes over
the training split, the budget falls as the lattice grows, from \baseEpochsLsixteen{} epochs at the two
smallest sizes to \baseEpochsLtwofiftysix{} at the largest. Counted in node labels, it is roughly
constant: from $L=32$ on, every network at the standard budget sees between \baseNodeLabelsLow{} and \baseNodeLabelsHigh{}
billion node labels in training, and \baseNodeLabelsLsixteen{} billion at $L=16$. The final model is scored on all \nTestGraphs{} official test graphs and on all
\nTestGraphs{} validation graphs. The score of a model is the base-ten logarithm of the mean over
test graphs of each graph's mean squared error over its nodes, which is the benchmark's logMSE since
every graph has $L^2$ nodes. Three models with different seeds make each entry. At the standard
budget the dilated networks trained for \dcnnSmallHoursLow{} to \dcnnHoursHigh{} hours each on one
NVIDIA T4, with a peak memory of \dcnnMemLow{} to \dcnnMemHigh{} MB, and the operators for at most
\fnoHoursHigh{} hours each. Of the additional runs in \cref{tab:baselines} (the matched control,
twice the steps and the lower learning rate), the dilated network at twice the steps at $L=256$ took
longest, up to \dcnnLaterHoursHigh{} hours in total, split into two runs. Every run stayed inside the benchmark's cap of \benchHourCap{} hours of compute.

\begin{table}[tbp]
\centering
\small
\begin{tabular}{lccccccc}
\toprule
 & & & & \multicolumn{2}{c}{dilated CNN} & \multicolumn{2}{c}{small dilated CNN} \\
\cmidrule(lr){5-6}\cmidrule(lr){7-8}
Size & steps & batch & epochs & width & parameters & width & parameters \\
\midrule
$L=16$ & \baseStepsLsixteen{} & \baseBatchLsixteen{} & \baseEpochsLsixteen{} & \dcnnWidthLsixteen{} & \dcnnLsixteenParams{} & \dcnnSmallWidthLsixteen{} & \dcnnSmallLsixteenParams{} \\
$L=32$ & \baseStepsLthirtytwo{} & \baseBatchLthirtytwo{} & \baseEpochsLthirtytwo{} & \dcnnWidthLthirtytwo{} & \dcnnLthirtytwoParams{} & \dcnnSmallWidthLthirtytwo{} & \dcnnSmallLthirtytwoParams{} \\
$L=64$ & \baseStepsLsixtyfour{} & \baseBatchLsixtyfour{} & \baseEpochsLsixtyfour{} & \dcnnWidthLsixtyfour{} & \dcnnLsixtyfourParams{} & \dcnnSmallWidthLsixtyfour{} & \dcnnSmallLsixtyfourParams{} \\
$L=128$ & \baseStepsLonetwentyeight{} & \baseBatchLonetwentyeight{} & \baseEpochsLonetwentyeight{} & \dcnnWidthLonetwentyeight{} & \dcnnLonetwentyeightParams{} & \dcnnSmallWidthLonetwentyeight{} & \dcnnSmallLonetwentyeightParams{} \\
$L=256$ & \baseStepsLtwofiftysix{} & \baseBatchLtwofiftysix{} & \baseEpochsLtwofiftysix{} & \dcnnWidthLtwofiftysix{} & \dcnnLtwofiftysixParams{} & \dcnnSmallWidthLtwofiftysix{} & \dcnnSmallLtwofiftysixParams{} \\
\bottomrule
\end{tabular}
\caption{The step budget and batch at each size, which every network of \cref{sec:archs} uses at
the standard budget, the passes over the \nTrainGraphs{} training graphs that they make (steps times
batch over \nTrainGraphs{}), and the width and parameter count of the two dilated networks. Every model trains
with Adam at \baseLrRegistered{}, a cosine schedule to zero and the gradient norm clipped at \baseClip{},
except the eight-layer GCN, which uses a constant rate and no clipping. The polynomial spectral
networks have their own budget (\cref{sec:archs}).}
\label{tab:hyper}
\end{table}

\begin{widetable}[t]
\centering
\small
\setlength{\tabcolsep}{3pt}
\begin{tabular}{@{}lrccccc@{}}
\toprule
 & params & $L=16$ & $L=32$ & $L=64$ & $L=128$ & $L=256$ \\
\midrule
Dilated CNN & 2.0\,M & ${-}4.96 \pm 0.11$ & ${-}4.86 \pm 0.08$ & ${-}4.33 \pm 0.08$ & ${-}3.92 \pm 0.03$ & ${-}3.44 \pm 0.05$ \\
\quad every dilation one & 2.0\,M & -- & ${-}4.69 \pm 0.11$ & -- & ${-}2.88 \pm 0.004$ & -- \\
\quad at a quarter of the learning rate & 2.0\,M & -- & ${-}5.16 \pm 0.02$ & -- & ${-}4.36 \pm 0.21$ & -- \\
\quad at twice the steps & 2.0\,M & -- & -- & -- & ${-}4.38 \pm 0.05$ & ${-}3.99 \pm 0.09$ \\
\quad with the target's gate, one model & 2.0\,M & ${-}5.02$ & ${-}5.09$ & -- & -- & -- \\
Small dilated CNN & 0.3\,M & ${-}4.91 \pm 0.04$ & ${-}4.92 \pm 0.12$ & ${-}4.60 \pm 0.07$ & ${-}4.12 \pm 0.26$ & ${-}3.75 \pm 0.17$ \\
\midrule
FNO & 2.4\,M & ${-}4.50 \pm 0.09$ & ${-}3.64 \pm 0.18$ & ${-}1.38 \pm 0.06$ & ${-}0.16 \pm 0.04$ & $+0.49 \pm 0.03$ \\
\quad every mode kept & 2.0\,M & -- & ${-}4.47 \pm 0.09$ & ${-}3.89 \pm 0.04$ & -- & -- \\
\midrule
Eight-layer GCN, ours & -- & -- & -- & ${-}1.09$ & ${-}0.93$ & ${-}0.99$ \\
Best published, in-domain & -- & ${-}4.70$ & ${-}4.85$ & -- & -- & -- \\
Polynomial floor & -- & ${-}2.48$ & ${-}2.68$ & ${-}2.95$ & ${-}3.23$ & ${-}3.52$ \\
Larger strided filter, labels & -- & -- & ${-}6.75$ & ${-}7.41$ & ${-}7.60$ & ${-}7.58$ \\
\bottomrule
\end{tabular}

\caption{Standard architectures on the official test split, in logMSE at the harder coupling, lower
is better: the mean and standard deviation of three models trained at each size, with the parameter
count at \latticeLthirtytwoNodes{} nodes. The dilated network with every dilation one is the matched
control of \cref{sec:dcone}; the rows at a quarter of the learning rate and at twice the steps are the
checks of \cref{sec:lrcheck}; the row with the target's gate is the single model of
\cref{sec:distance} at each size. The eight-layer GCN is the mean of two models, trained on the same
data and for the same number of steps as the dilated network, but with a constant learning rate and no
clipping of the gradient. The
best published results are those of \citet{carotti2026ghr}. \Cref{tab:baselines} gives every model.}
\label{tab:models}
\end{widetable}

\begin{widefigure}[t]
\centering
\includegraphics[width=\linewidth]{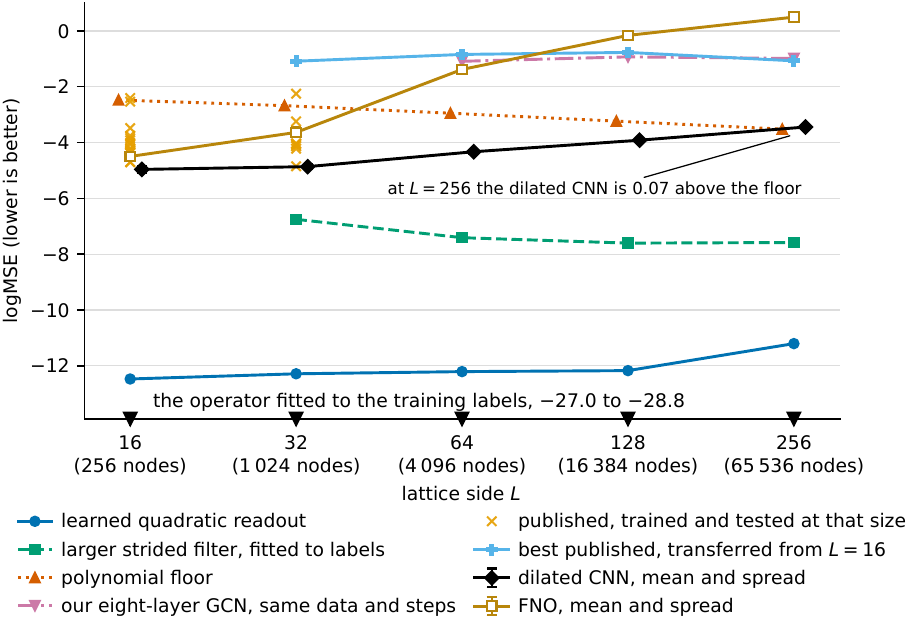}
\caption{Where models stand, at every lattice size; lower is better. The dilated network matches or beats
the published results and stays above the larger strided filter fitted to the training labels at every size where both
exist; the Fourier
neural operator falls from beside it at the smallest size to above the polynomial floor from
\fnoAboveFloorFromNodes{} nodes. The learned quadratic readout sits \gapBenchToReadoutThirtyTwo{} to \gapBenchToReadoutSixteen{} orders of magnitude below the best
published in-domain results, and the operator fitted to the training labels is marked at the bottom
edge. The published transfer line gives the best published result at each size for a model trained at
\benchNodesLow{} nodes. Our eight-layer GCN had the same data and number of steps as the
dilated network, with a constant learning rate, no clipping and two models per size.}
\label{fig:stand}
\end{widefigure}

\section{Results}
\label{sec:models}

Do trained networks bear out the two requirements of \cref{sec:floor}? We trained standard
architectures on the official splits at every size and, at fewer sizes, a matched control,
polynomial spectral networks and graph networks, under the protocol of \cref{sec:setup}, and the
dilated network again at the benchmark's easier coupling.

\Cref{tab:models} and \cref{fig:stand} give the results. The dilated network passes the polynomial
floor at every size but the largest, from \dcnnLsixteenRound{} at \benchNodesLow{}
nodes to \dcnnLonetwentyeightRound{} at \latticeLonetwentyeightNodes{}, and at \benchNodesHigh{}
nodes it lands \dcnnAboveFloorTwoFiftySix{} above the floor at the standard budget and \dcnnTwiceBelowFloorLtwofiftysix{} below it
at twice the steps. The benchmark's plain convolutional networks fell behind the best graph networks
\citep[App.~E.5]{mathys2026lrim}. The dilated network beats the best published result, that of
\citet{carotti2026ghr}, by \dcnnOverPublishedSixteen{} at \benchNodesLow{} nodes and matches it at
\latticeLthirtytwoNodes{}, where the difference, \dcnnOverPublishedThirtyTwo{}, is within the spread of its three models. It beats our eight-layer GCN, on the same data and
number of steps, by \dcnnOverGcnLow{} to \dcnnOverGcnHigh{} orders of magnitude at the three larger sizes.
The small version, with \dcnnSmallFractionWord{} of the parameters, does as well or better, down to
\dcnnSmallLtwofiftysixRound{} at \benchNodesHigh{} nodes, below the floor. The larger network's
recipe was not tuned, and a quarter of the learning rate or twice the steps improves it
(\cref{sec:lrcheck}).

The Fourier neural operator reaches the whole lattice in every layer, yet it trails the dilated
network at every size, least at the smallest, where it keeps every mode the lattice has. As the
lattice grows it slides from \fnoLsixteenRound{} to \fnoLtwofiftysixRound{}, worse than a polynomial
filter can do from \fnoAboveFloorFromNodes{} nodes on. With its modes fixed, the operator keeps a
shrinking share of the frequencies as the lattice grows, and the coupling's short range lives in the
high ones. Keeping every mode, at a smaller width that keeps the parameter count, recovers about
\fnoAllModesShareWord{} of its gap to the dilated network at \latticeLthirtytwoNodes{} nodes and about
\fnoAllModesShareLsixtyfourWord{} at \latticeLsixtyfourNodes{} nodes, a size we chose in advance to
test this account. With every mode kept, the operator has global reach, full resolution and no
strides, and it trails the dilated network by \fnoAllModesBehindDcnn{} at both sizes. Truncation
therefore accounts for most of the operator's gap to the dilated network.

Truncation does not set the operator's level, however. The linear part of a Fourier layer is a free
multiplier on the kept modes plus a pointwise channel map, which is the same matrix at every
frequency, so the symbol of the linearized network is one constant outside the box of kept modes. For
a readout of the target's form with such a symbol, the error is at least the $S$-weighted variance of
$\hat{J}$ outside the box, by the argument of \cref{prop:floor}. We computed this floor from the
official training split before any model was scored: at \fnoModes{} modes per axis it is
\fnoModeFloorLthirtytwoTwo{}, \fnoModeFloorLsixtyfourTwo{}, \fnoModeFloorLonetwentyeightTwo{} and
\fnoModeFloorLtwofiftysixTwo{} from $L=32$ to $L=256$, and there is none at $L=16$, where every mode is
kept. Like \cref{prop:floor} it binds the linear class only. We had expected the trained operator to
lie between \expFnoModeBandBelow{} below and \expFnoModeBandAbove{} above it; it lies below it at
every size, by \fnoBelowModeFloorThirtyTwo{} at $L=32$. The trained operator is not of the target's
form, since it must build the gate itself and its pointwise path carries the spin at full resolution,
so the floor of its linearized form does not bound it.

The small operator, with about \fnoSmallFractionWord{} of the parameters, is worse than the large one
at every size, by \fnoSmallBehindLow{} to \fnoSmallBehindHigh{}. Adding the reference code's two grid
channels makes the operator worse by \fnoGridWorseLthirtytwo{} at \latticeLthirtytwoNodes{} nodes.

\subsection{A matched control without strides}
\label{sec:dcone}

Do the strides do the work, or the depth, width and residual blocks around them? The control is the
same network at the same width, depth and budget with every dilation set to one, trained at
\latticeLthirtytwoNodes{} and \latticeLonetwentyeightNodes{} nodes, three models at each size. At
\latticeLthirtytwoNodes{} nodes its stack of $3\times3$ convolutions still covers the whole torus,
and it scores \dcOneGainLthirtytwo{} worse than with strides, where we had expected at least
\expDcOneMinGainLthirtytwo{}, and below the polynomial floor. At
\latticeLonetwentyeightNodes{} nodes the same stack reaches \dcOneReachLonetwentyeight{} sites along
each axis, short of half the lattice. There the control scores \dcOneLonetwentyeightRound{},
\dcOneGainLonetwentyeight{} worse than with strides and
\dcOneAbovePolyFloorLonetwentyeight{} above the polynomial floor. A filter of the target's form supported on that box leaves \dcOneBoxFloorLonetwentyeight{},
and the control lands \dcOneAboveBoxFloorLonetwentyeight{} above it. This is a reference rather than a bound, since a nonlinear
network could use the correlations between spins (\cref{sec:benchmark}) to infer some of what lies
beyond its reach.
Strides let the dilated network cover the lattice with a depth that grows only as $\log L$.

\subsection{Polynomial spectral networks against the floor}
\label{sec:spectral}

Does the floor predict spectral networks built on the class it addresses? We fitted or trained adaptations of two
published spectral architectures at \benchNodesLow{}, \latticeLthirtytwoNodes{} and \latticeLsixtyfourNodes{} nodes
(\cref{sec:archs}), each with $K=8$ rounds of propagation and with $K=L$ (\cref{tab:spectral}). The
gated linear filter, GPR-GNN's propagation inside the target's gate, is a readout of the form that
\cref{prop:floor} covers, and fitted to the training labels it lands within \specGatedMaxDiff{} of
the closed-form best polynomial of its degree at every size and both depths, so the closed form gives
its test score. The SIGN readout feeds the same channels to a nonlinear network that the proposition does not
cover. It lands within a tenth of the closed form of its depth at every size, \specSignBelowLow{} to
\specSignBelowHigh{} below it, and at $K=L$ it stays \specSignAboveFloorLow{} to
\specSignAboveFloorHigh{} above the floor. On these lattices the floor and the degree ladder of
\cref{tab:closed} therefore predict both adaptations to within a tenth of a decade,
and propagating as far as the diameter of the lattice does not bring it to the floor.

\begin{table}[tbp]
\centering
\small
\begin{tabular}{lccc}
\toprule
 & $L=16$ & $L=32$ & $L=64$ \\
\midrule
\multicolumn{4}{l}{\emph{Eight rounds of propagation, $K=8$}} \\
Gated linear filter, least squares & ${-}1.632$ & ${-}1.179$ & ${-}1.010$ \\
SIGN, mean of three models & ${-}1.691$ & ${-}1.211$ & ${-}1.044$ \\
Best polynomial of degree eight (closed form) & ${-}1.636$ & ${-}1.179$ & ${-}1.012$ \\
\midrule
\multicolumn{4}{l}{\emph{As many rounds as the side of the lattice, $K=L$}} \\
Gated linear filter, least squares & ${-}1.968$ & ${-}1.923$ & ${-}1.907$ \\
SIGN, mean of three models & ${-}2.053$ & ${-}1.978$ & ${-}1.954$ \\
Best polynomial of degree $L$ (closed form) & ${-}1.973$ & ${-}1.921$ & ${-}1.907$ \\
\midrule
Floor for any polynomial filter (closed form) & ${-}2.475$ & ${-}2.678$ & ${-}2.949$ \\
\bottomrule
\end{tabular}

\caption{Polynomial spectral networks on the official test split, in logMSE at the harder coupling,
lower is better, with the closed forms of \cref{tab:closed} as references. $K$ is the number of
rounds of propagation, the highest power of the adjacency the channels reach. The gated linear filter
is fitted by least squares and has no seeds; each SIGN entry is the mean of three models, whose
standard deviation is below \specSignSdCeil{} in every cell.}
\label{tab:spectral}
\end{table}

\subsection{Graph networks under the same protocol}
\label{sec:graphnets}

Do graph networks that cover the lattice and are not bound by the floor close the gap to the dilated network? At their
published depths the gated graph network \citep{bresson2017gated} and its variants cover the lattice
at \benchNodesLow{} and \latticeLthirtytwoNodes{} nodes, and their gates make them nonlinear, so the
floor does not bind them. We trained four with about two million parameters under the protocol above
(\cref{tab:graphmodels}, which gives each network's depth and width): the gated network as the benchmark configures it, the same network with a
separate weight for each edge direction, which frees even its linear part from the floor, that
network with edges at one more stride in each layer, cycling through $2, 4, \ldots, L/2$, and the
gated network with global attention \citep{rampasek2022gps}. We trained each three times at
\benchNodesLow{} nodes and the first three once at \latticeLthirtytwoNodes{} nodes. At \benchNodesLow{} nodes all four score below the polynomial floor, the weakest by
\gnnWorstOverPolyFloorLsixteen{}, and all four trail the dilated network, the best by
\gnnBestBehindDcnnLsixteen{}. The separate weights per direction gain only \gnnDirGainLsixteen{}, and
at \latticeLthirtytwoNodes{} nodes almost nothing, where we had expected at least \expDirMinGain{}. At this budget a stride per
layer or global attention gains about \gnnShortcutGainLsixteen{} over the gated network. At \latticeLthirtytwoNodes{} nodes
the stride gains \gnnCycGainLthirtytwo{}, a difference between two single models that lies within
the strided network's seed spread at \benchNodesLow{} nodes.

\begin{table}[tbp]
\centering
\small
\setlength{\tabcolsep}{3pt}
\resizebox{\linewidth}{!}{\begin{tabular}{@{}llcrrcccccc@{}}
\toprule
Model & $L$ & training & layers $\times$ width & parameters & \multicolumn{3}{c}{each model, test} & mean & s.d. & published \\
\midrule
GatedGCN & 16 & standard & 20 $\times$ 104 & 1\,995\,553 & ${-}2.906$ & ${-}2.970$ & ${-}2.977$ & ${-}2.951$ & $0.039$ & ${-}3.844$ \\
 & 16 & $4\times$ steps & 20 $\times$ 104 & 1\,995\,553 & ${-}3.747$ & ${-}3.975$ & ${-}3.736$ & ${-}3.819$ & $0.135$ & ${-}3.844$ \\
 & 32 & standard & 32 $\times$ 84 & 2\,087\,905 & ${-}3.249$ & -- & -- & ${-}3.249$ & -- & ${-}4.087$ \\
\quad per-direction & 16 & standard & 20 $\times$ 80 & 1\,965\,121 & ${-}2.989$ & ${-}3.035$ & ${-}3.042$ & ${-}3.022$ & $0.029$ & -- \\
 & 32 & standard & 32 $\times$ 64 & 2\,019\,585 & ${-}3.283$ & -- & -- & ${-}3.283$ & -- & -- \\
\quad per-direction, strided & 16 & standard & 20 $\times$ 64 & 1\,929\,473 & ${-}3.620$ & ${-}3.372$ & ${-}3.376$ & ${-}3.456$ & $0.142$ & -- \\
 & 16 & $4\times$ steps & 20 $\times$ 64 & 1\,929\,473 & ${-}3.847$ & ${-}4.061$ & ${-}4.047$ & ${-}3.985$ & $0.120$ & -- \\
 & 32 & standard & 32 $\times$ 52 & 2\,046\,305 & ${-}3.363$ & -- & -- & ${-}3.363$ & -- & -- \\
GPS with RWSE & 16 & standard & 16 $\times$ 96 & 1\,960\,893 & ${-}3.647$ & ${-}3.337$ & ${-}3.367$ & ${-}3.451$ & $0.171$ & ${-}4.011$ \\
\bottomrule
\end{tabular}
}
\caption{Graph networks under our protocol, in test logMSE at the harder coupling, lower is
better. Each runs at the depth its benchmark configuration uses, with the width set for about two
million parameters. The standard rows use the dilated network's optimizer and step budget; the rows
at four times the steps use half its learning rate. The indented rows are GatedGCN with a separate
weight for each edge direction (per-direction), and that network with edges at one more stride in
each layer, cycling through $2, 4, \ldots, L/2$ (strided). Rows at
$L=32$ are single models. The published column is the in-domain result for the same
architecture from the benchmark's main table, not the diagonal of its transfer table.}
\label{tab:graphmodels}
\end{table}

These networks also sit above their published numbers, the gated network at \gnnGatedLsixteen{}
against \gnnGatedLsixteenPublished{}. Our protocol gives every model the dilated network's budget,
and the gated networks need more. We trained the gated network and its strided version three times
each for four times the steps at half the learning rate. The gated network reaches
\gnnGatedLongLsixteen{}, \gnnGatedLongShortOfPublished{} short of its published value, and the
network with strides \gnnCycLongLsixteen{}. Their difference, \gnnCycLongGainOverGatedLongShown{}, is
close to the spread between seeds, about \gnnLongSdRound{}, so the stride's lead of about
\gnnShortcutGainLsixteen{} at the standard budget largely vanishes. We suspect that where depth
already covers the lattice, strides mainly speed up training. The closer of the two stays
\gnnCycLongBehindDcnnLsixteen{} behind the dilated network, which had a quarter of the steps and the
higher rate. At this size, covering the lattice and being free of the floor did not close the gap
to the dilated network, which also weighs the lattice's diagonal neighbors separately.

\subsection{The learning-rate check}
\label{sec:lrcheck}

Would a lower learning rate have served the networks better? The dilated network and the operator were each trained once more
at $L=32$, with one seed, at two lower learning rates, and read on the validation split
(\cref{tab:lr}), which is the split the check was planned on. The operator does worse at both, by
\lrFnoHighDiff{} and \lrFnoLowDiff{}. For the operator at \baseLrHigh{} only the validation score is
available, so the table gives validation scores only. The dilated network does better at both, by \lrDcnnHighAbs{} and
\lrDcnnLowAbs{}. We then trained it at \baseLrLow{} as a family of three models at
\latticeLthirtytwoNodes{} and \latticeLonetwentyeightNodes{} nodes, and at twice the steps at the two
largest sizes, each planned before the runs. At a quarter of the rate it reaches
\dcnnSlowRateTwoLthirtytwo{} at \latticeLthirtytwoNodes{} nodes and \dcnnSlowRateTwoLonetwentyeight{}
at \latticeLonetwentyeightNodes{}, against \dcnnTwoLthirtytwo{} and \dcnnTwoLonetwentyeight{} at the
standard rate. At twice the steps it reaches \dcnnTwiceTwoLonetwentyeight{} at
\latticeLonetwentyeightNodes{} nodes and \dcnnTwiceTwoLtwofiftysix{} at \benchNodesHigh{}, against
\dcnnTwoLonetwentyeight{} and \dcnnTwoLtwofiftysix{} at the standard budget (\cref{tab:models}). Both
help, by similar amounts at \latticeLonetwentyeightNodes{} nodes; we did not test whether they add.

\subsection{The distance that remains}
\label{sec:distance}

How far are these models from the operator itself? From \latticeLthirtytwoNodes{} nodes on, the
dilated network at its standard budget stays \dcnnToFilterLow{} to \dcnnToFilterHigh{} orders of
magnitude worse than the larger strided filter fitted to the training labels, and at the step budgets
we used the gap grows with size, because the network loses accuracy as the lattice grows while the
filter does not. The budget per
size falls as the lattice grows, and part of the loss is budget: at twice the steps the dilated
network recovers about \dcnnTwiceDeclineShareWord{} of what it loses between
\latticeLthirtytwoNodes{} and \benchNodesHigh{} nodes. The learned readout of \cref{eq:readout}, which has the target's form, is
\gapDcnnToReadoutThirtyTwo{} orders of magnitude below the dilated network at
\latticeLthirtytwoNodes{} nodes.

Does the product with the node's own spin account for part of that distance? The readout carries
the target's gate $s_i$, and no network does. We gave the dilated network that gate: its output at
node $i$ is a learned constant plus $s_i$ times the network's output, so what remains for it to learn
is linear in the spin field. Trained once at each of the two smallest sizes under the same protocol,
it scores \gatedDcnnLsixteen{} at \benchNodesLow{} nodes, \gatedDcnnGainShownLsixteen{} better than the
dilated network and within its seed spread, and \gatedDcnnLthirtytwo{} at
\latticeLthirtytwoNodes{} nodes, \gatedDcnnGainShownLthirtytwo{} better, about
\gatedDcnnGainSpreadsWord{} times the seed spread there but short of the gain of at least \expGatedDcnnMinGain{} we had
expected. It stays \gatedDcnnToFilterThirtyTwo{} orders of magnitude above the larger
strided filter fitted to the same labels. For the dilated network at these two sizes, the product
with the spin explains little of the distance; what remains lies in learning the coupling, and we
cannot say why training stops where it does.

\subsection{The easier coupling}
\label{sec:secondcoupling}

Does the analysis hold when the coupling decays faster? LRIM publishes every size at the exponent
\benchSigmaEasy{} as well, and we repeated the model-free analysis there and trained the dilated
network with the same recipe and seeds (\cref{tab:secondcoupling}; \cref{app:secondcoupling} gives
the details). Least squares on the training labels again recovers the operator, to between
\secondExactTiedShallowRound{} and \secondExactTiedDeepRound{}. The polynomial floor lies between
\secondFloorShallowRound{} and \secondFloorDeepRound{}, close to where it lies at the harder coupling.
The fifty-coefficient strided filter fitted to the training labels reaches
\secondFiftyCoefShallowRound{} to \secondFiftyCoefDeepRound{}; at \latticeLthirtytwoNodes{} nodes that
is \secondFloorToFiftyCoefLthirtytwo{} orders of magnitude below the floor and
\secondFiftyCoefBelowPublishedLthirtytwo{} below the best published network, \secondPublishedModel{}.
The dilated network beats \secondPublishedModel{} by \secondDcmBelowPublishedLthirtytwo{} at
\latticeLthirtytwoNodes{} nodes and matches it at \benchNodesLow{} nodes, where the difference,
\secondDcmBelowPublishedLsixteen{}, is less than the standard deviation of its three models. It stays \secondDcmToFiftyCoefLow{} to
\secondDcmToFiftyCoefHigh{} orders of magnitude above the fifty-coefficient filter at every size, and
further above the larger one.

The dilated network does better against the floor here than at the harder coupling. It passes the
floor at every size, by \secondDcmBelowFloorLow{} to \secondDcmBelowFloorHigh{}, and it loses less as
the lattice grows, going from \secondDcmLsixteenRound{} at \benchNodesLow{} nodes to
\secondDcmLtwofiftysixRound{} at \benchNodesHigh{}, against \dcnnLsixteenRound{} to
\dcnnLtwofiftysixRound{} at the harder coupling. A faster decay puts more of the coupling's weight
near the node, which is consistent with this, though we did not test it.

\section{Discussion}
\label{sec:discussion}

\subsection{What this means for long-range benchmarks}
\label{sec:implications}

What reference points would help readers of LRIM and of benchmarks built like it? LRIM's authors note
that methods aligned with the known data generation could score better, and they hold general graph
methods to be the more informative test \citep[App.~E.5]{mathys2026lrim}. The references below are
meant to sit beside that test.

Report the closed-form reference whenever the target is a known operator. On LRIM a least-squares
fit to the training labels beside the leaderboard would show each model's distance from a known
answer.

Report the floor for a model class when one can be computed. The projection argument behind
\cref{prop:floor} needs only spins of plus or minus one, a target that is the spin times a translation-invariant
filter of the spin field, and a torus or another abelian Cayley graph, and it tells a reader
which readouts of that form are excluded before any is trained.

Report a model that knows the domain as a reference line. On a lattice the standard one is a dilated
convolutional network \citep{yu2016dilated, yu2017drn}, which weighs separately the offsets the
adjacency lumps together and reaches the whole lattice. It uses the coordinates that general graph
methods do without, so it marks how much of the distance that knowledge closes rather than competing
with them. On general graphs, multi-hop or dilated edges \citep{li2019deepgcns} and direction-aware
aggregation \citep{beaini2021dgn} play that role.

Complement size transfer with in-domain results. The larger sizes of LRIM serve as a test of
transfer from the smallest size \citep[Table~3]{mathys2026lrim}, and in-domain results at the same
sizes separate what an architecture can express from how well it transfers. On this task the dilated
network and a Fourier operator that reaches the whole lattice in every layer come apart further as
the lattice grows, which a single size would not show.

\subsection{Limitations}
\label{sec:limitations}

Where do these conclusions stop? The instruments are built for lattices.
The strided filter, the dilated network and the quadratic readout all use the lattice's shift
structure, and none of them runs as written on a general graph. What transfers is the method of
reading the target as an operator and measuring each model against the closed-form reference and the
floor of its class.

The floor binds readouts of the target's form whose filter is a function of the adjacency's
spectrum. Gated and attention networks score below it, so it does not say where they stop, and we
trained them under our protocol only at the two smallest sizes, mostly with one budget that leaves them
short of their published numbers. At the three larger sizes the in-domain comparisons for the dilated
network are therefore the Fourier operator and the eight-layer GCN, and the GCN was trained with a
constant learning rate and no clipping, so a dilated network leads at every size only among the
models we trained. Our account of the Fourier operator, that discarding high modes
accounts for most of its gap to the dilated network, was formed after its results at one size and
tested at a second size chosen in advance; it does not explain the operator's level, which lies below
the floor of its linearized form.

Our baselines follow one fixed budget per size, not the benchmark's epochs-and-patience protocol,
and were not tuned; a lower learning rate or twice the steps improves the dilated network,
and we did not test whether a longer budget would recover the rest of its loss with size. At the easier coupling
we trained only the dilated network, so the comparisons with the Fourier operator and the graph networks hold at
the harder coupling alone.

We cannot say why the dilated network stops where it does, \gapDcnnToReadoutThirtyTwo{}
orders of magnitude above the learned readout of \cref{eq:readout} at \latticeLthirtytwoNodes{} nodes. Giving it
the target's gate narrows this by only \gatedDcnnGainLthirtytwo{} in one model at that size
(\cref{sec:distance}), so the product with the node's spin explains little of it there, and we did
not test whether a longer budget or another optimizer would close the rest.

\section{Conclusion}
\label{sec:conclusion}

LRIM's target is one quadratic convolution with a known kernel. A filter of the target's form must weigh separately the offsets the adjacency
lumps together, which no polynomial in the adjacency does, and must resolve the coupling out to the
whole lattice. Spectral networks adapted from published ones land where the best polynomial of their degree does. A
dilated convolutional network with both properties matches or beats the best published graph networks
at the two sizes they were trained at, at both couplings. Gated graph networks, whose depth covers the lattice and which the floor does not bind,
still trail it at the smallest size, so these properties do not decide a trained network's score. Every trained network trails
the strided filters fitted to the labels by orders of magnitude.

\bibliographystyle{elsarticle-harv}
\bibliography{refs}

\appendix
\makeatletter
\gdef\thesection{\@Alph\c@section}
\gdef\thetable{\arabic{table}}
\gdef\thefigure{\arabic{figure}}
\makeatother
\sloppy 
\section{Checks of the operator identity and of the polynomial fits}
\label{app:proofs}

\subsection{The operator identity}

The generator draws a spin configuration $s \in \{-1,+1\}^N$ on the $L \times L$ torus and labels
node $i$ with its interaction energy under the coupling $J(r) = r^{-(2+\sigma)}$, where $r$ is the
shortest distance on the torus and the self term is excluded. Summing the pairwise energies that
touch node $i$ gives \cref{eq:target}. We did not take the identity from the description; we
checked it numerically, in two ways. On \analyticCheckGraphs{} official graphs at $L=32$, with the
labels read without conversion to single precision, $s \odot (J \circledast s)$ evaluated in double
precision matches them to a logMSE of \exactAnalyticCheckRound{}, where predicting zero scores
\analyticCheckPowerRound{}. At every size, on \officialResidualGraphs{} graphs of the official training split, with the labels
kept in single precision as they are when a model trains on them, the largest residual lies below
\officialRelErrRound{}, which is the rounding of single precision. That
residual is the maximum absolute difference divided by the largest label, not a per-element relative
error, and the scale relating prediction and label is one, to six decimal places. The same check
confirms the row-major ordering of the published node features on the first graph at each size. At
$L=256$ it ran on a copy of the data that holds the last \readoutTrainSubsetLtwofiftysix{} training
configurations and the full validation and test splits. The learned readout at $L=256$ was trained on
that copy (\cref{sec:archs}); no network was.

\subsection{The operator fitted to the labels}

The fit of \cref{sec:benchmark} needs no design matrix. For spins of $\pm 1$, \cref{eq:target} is
equivalent to $s_i y_i = (J \circledast s)_i$, so least squares for a translation-invariant kernel
separates by Fourier mode and has a closed-form solution at each one; the fit with the symmetries of
the square holds the kernel's transform constant on each orbit of those symmetries and solves for
one value per orbit. It reads the labels of all \nTrainGraphs{} training configurations in double
precision from the official files, uses the coupling nowhere and is scored on all \nTestGraphs{} test
graphs. The coupling itself, scored the same way, reaches \exactFitJLsixteenRound{} to
\exactFitJLtwofiftysixRound{} from the smallest size to the largest, and \cref{tab:stand} gives the fit
beside it at every size. An earlier implementation of the same
fit, whose code was not kept, gave values \exactFitOldGapLow{} to \exactFitOldGapHigh{} higher at
the three sizes where it was run; the values we report are those of the implementation we can rerun.

\subsection{The basis for the polynomial fits}

We solved the polynomial fits of \cref{sec:floor} in the basis of polynomials orthonormal for the weights on the levels, built by
Lanczos with full re-orthogonalization, which stays accurate to numerical precision up to degree one less than the number of
levels; a Chebyshev design matrix solved by singular value decomposition loses rank at degree 128
on the \polyLevelsLthirtytwo{} levels of $L=32$ and understates the fit there. The monomial basis
was run beside it as a numerical control. Its design matrix loses rank from degree 64 on, and with it
the fit appears to saturate near \priorPolySaturation{} with degree at $L=32$; that apparent saturation
is an artifact of the basis, not a property of polynomial filters.

\section{The numbers in full}
\label{app:perscale}

\Cref{tab:stand} places everything we could put on one axis: the closed forms, the learned readouts,
the standard baselines, the published results of both kinds and our eight-layer GCN. The baselines
are given model by model in \cref{tab:baselines}, and the closed forms in \cref{tab:closed}. The first block of
\cref{tab:stand} is closed form, with predicting zero as a trivial reference. Its rows fitted to the
labels are scored on the test split; its rows fitted to the coupling, the floor and predicting zero
are computed with the spin spectrum of \polySpectrumGraphs{} training configurations.
The second block is a readout of the target's own functional form, trained by gradient descent. The
third holds the standard architectures, trained and tested at each size. The fourth and fifth are
published results, kept apart because a model trained at one size and tested at another answers a
different question. The last is our eight-layer GCN, trained on the same data and for the same
number of steps as the dilated networks, with a constant learning rate and no clipping. \Cref{tab:lr} gives the learning-rate check of \cref{sec:lrcheck}.

\begin{table}[htbp]
\centering
\scriptsize
\setlength{\tabcolsep}{2.6pt}
\resizebox{\linewidth}{!}{\begin{tabular}{lccccc>{\raggedright\arraybackslash}p{0.14\linewidth}}
\toprule
Model or bound & $L=16$ & $L=32$ & $L=64$ & $L=128$ & $L=256$ & note \\
\midrule
\multicolumn{7}{l}{\emph{Closed forms: the operator solved exactly, and the floors}} \\
Least squares on the coupling, 152 coefficients & -- & ${-}27.410$ & -- & -- & -- & fitted to the kernel, not the labels \\
Least squares on the labels, symmetric kernel & ${-}28.849$ & ${-}28.709$ & ${-}28.076$ & ${-}27.426$ & ${-}27.034$ & the square's symmetries \\
Least squares on the labels, free kernel & ${-}28.850$ & ${-}28.709$ & ${-}28.077$ & ${-}27.434$ & ${-}27.043$ & $L^2$ coefficients \\
The coupling $J$, applied to the spins & ${-}28.947$ & ${-}28.624$ & ${-}28.059$ & ${-}27.365$ & ${-}26.998$ & in double precision \\
Larger strided filter & -- & ${-}6.752$ & ${-}7.410$ & ${-}7.605$ & ${-}7.583$ & 83 to 101 coefficients, fitted to the labels \\
Fifty-coefficient strided filter & -- & ${-}6.567$ & ${-}6.972$ & -- & -- & 49 and 55 coefficients, rank 43 and 48, fitted to the coupling \\
\quad top level at half the step & -- & ${-}6.625$ & ${-}7.003$ & -- & -- & top taps at a quarter of the lattice, fitted to the coupling \\
\quad learned from the training labels & -- & ${-}6.566$ & ${-}6.973$ & -- & -- & the same basis, scored on the test split \\
Best polynomial filter in $A$ & ${-}2.475$ & ${-}2.458$ & ${-}1.977$ & ${-}1.898$ & ${-}1.885$ & degree 128, fitted to the coupling \\
\quad floor for any such filter & ${-}2.475$ & ${-}2.678$ & ${-}2.949$ & ${-}3.233$ & ${-}3.518$ & any degree; exact for the training spectrum \\
Predicting zero at every node & $+1.465$ & $+1.286$ & $+1.221$ & $+1.173$ & $+1.047$ & the label's power \\
\midrule
\multicolumn{7}{l}{\emph{A learned quadratic convolution readout, trained by gradient descent}} \\
Free Fourier multiplier & ${-}12.473$ & ${-}12.287$ & ${-}12.209$ & ${-}12.174$ & ${-}11.206$ & 257 to 65\,537 parameters \\
Larger strided filter & ${-}10.496$ & ${-}6.752$ & ${-}7.409$ & ${-}7.603$ & ${-}7.579$ & 57 to 102 parameters \\
\quad nine gates in place of one & -- & ${-}12.496$ & -- & -- & -- & 9\,217 parameters, \cref{app:protocol} \\
\midrule
\multicolumn{7}{l}{\emph{Standard architectures, trained and tested at each size (three models each)}} \\
Dilated CNN & ${-}4.963 \pm 0.112$ & ${-}4.864 \pm 0.078$ & ${-}4.330 \pm 0.078$ & ${-}3.916 \pm 0.028$ & ${-}3.445 \pm 0.049$ & \cref{app:baselines} \\
\quad every dilation one & -- & ${-}4.687 \pm 0.111$ & -- & ${-}2.881 \pm 0.004$ & -- & \cref{app:baselines} \\
Small dilated CNN & ${-}4.910 \pm 0.039$ & ${-}4.920 \pm 0.122$ & ${-}4.600 \pm 0.073$ & ${-}4.117 \pm 0.261$ & ${-}3.746 \pm 0.166$ & \cref{app:baselines} \\
FNO, published configuration & ${-}4.498 \pm 0.094$ & ${-}3.637 \pm 0.180$ & ${-}1.381 \pm 0.064$ & ${-}0.161 \pm 0.044$ & $+0.494 \pm 0.026$ & \cref{app:baselines} \\
\quad every mode kept & -- & ${-}4.467 \pm 0.087$ & ${-}3.890 \pm 0.040$ & -- & -- & \cref{app:baselines} \\
\quad with grid coordinates & -- & ${-}3.378 \pm 0.099$ & -- & -- & -- & \cref{app:baselines} \\
Small FNO & ${-}4.053 \pm 0.099$ & ${-}2.791 \pm 0.109$ & ${-}0.542 \pm 0.033$ & $+0.212 \pm 0.083$ & $+0.645 \pm 0.019$ & \cref{app:baselines} \\
\midrule
\multicolumn{7}{l}{\emph{Published, trained and tested at the same size. Nothing is published above $L=32$; $\ell$ is the number of layers}} \\
GIN & ${-}2.533 \pm 0.313$ & ${-}2.249 \pm 0.135$ & -- & -- & -- & $\mathcal{O}(\ell \cdot E)$ \\
GatedGCN & ${-}3.844 \pm 0.055$ & ${-}4.087 \pm 0.238$ & -- & -- & -- & $\mathcal{O}(\ell \cdot E)$ \\
GatedGCN-VNG & ${-}4.068 \pm 0.131$ & ${-}3.243 \pm 0.177$ & -- & -- & -- & $\mathcal{O}(\ell \cdot E + \ell \cdot N)$ \\
GPS-Base & ${-}4.211 \pm 0.155$ & ${-}4.044 \pm 0.122$ & -- & -- & -- & $\mathcal{O}(\ell \cdot N^{2})$ \\
GPS-RWSE & ${-}4.011 \pm 0.129$ & ${-}4.134 \pm 0.075$ & -- & -- & -- & $\mathcal{O}(\ell \cdot N^{2})$ \\
GPS-LapPE & ${-}4.334 \pm 0.065$ & ${-}4.032 \pm 0.092$ & -- & -- & -- & $\mathcal{O}(\ell \cdot N^{2})$ \\
GHR-GatedGCN & ${-}4.701 \pm 0.031$ & ${-}4.852 \pm 0.013$ & -- & -- & -- & \citet{carotti2026ghr} \\
GHR-GIN & ${-}4.378 \pm 0.005$ & ${-}4.220 \pm 0.011$ & -- & -- & -- & \citet{carotti2026ghr} \\
LGSM & ${-}4.284 \pm 0.133$ & -- & -- & -- & -- & \citet{mathys2026lgsm} \\
HOPPER & ${-}3.485 \pm 0.056$ & -- & -- & -- & -- & \citet{herath2026hopper} \\
\midrule
\multicolumn{7}{l}{\emph{Published, trained at one size and tested at another. These are not comparable with the rows above}} \\
GIN\,$\dagger$ & ${-}2.406 \pm 0.148$$^{\ddagger}$ & ${-}1.043 \pm 0.051$ & ${-}0.774 \pm 0.042$ & ${-}0.703 \pm 0.041$ & ${-}0.903 \pm 0.047$ & trained at $L=16$ \\
GatedGCN\,$\dagger$ & ${-}3.919 \pm 0.223$$^{\ddagger}$ & ${-}1.050 \pm 0.004$ & ${-}0.781 \pm 0.002$ & ${-}0.708 \pm 0.003$ & ${-}0.952 \pm 0.005$ & trained at $L=16$ \\
GatedGCN-VNG\,$\dagger$ & ${-}3.756 \pm 0.063$$^{\ddagger}$ & ${-}1.054 \pm 0.006$ & ${-}0.788 \pm 0.004$ & ${-}0.716 \pm 0.005$ & ${-}0.968 \pm 0.008$ & trained at $L=16$ \\
GPS-Base\,$\dagger$ & ${-}4.340 \pm 0.101$$^{\ddagger}$ & ${-}1.057 \pm 0.000$ & ${-}0.790 \pm 0.001$ & ${-}0.719 \pm 0.001$ & \emph{n/a} & trained at $L=16$; out of memory at $L=256$ \\
GPS-RWSE\,$\dagger$ & ${-}4.345 \pm 0.065$$^{\ddagger}$ & ${-}1.057 \pm 0.005$ & ${-}0.790 \pm 0.004$ & ${-}0.716 \pm 0.002$ & \emph{n/a} & trained at $L=16$; out of memory at $L=256$ \\
GPS-LapPE\,$\dagger$ & ${-}4.248 \pm 0.110$$^{\ddagger}$ & ${-}1.053 \pm 0.006$ & ${-}0.785 \pm 0.005$ & ${-}0.716 \pm 0.005$ & \emph{n/a} & trained at $L=16$; out of memory at $L=256$ \\
GHR-GatedGCN (from $16$)\,$\dagger$ & -- & ${-}1.086 \pm 0.014$ & ${-}0.825 \pm 0.023$ & ${-}0.771 \pm 0.001$ & ${-}1.073 \pm 0.003$ & trained at $L=16$ \\
GHR-GIN (from $16$)\,$\dagger$ & -- & ${-}1.085 \pm 0.021$ & ${-}0.843 \pm 0.030$ & ${-}0.770 \pm 0.011$ & ${-}1.066 \pm 0.033$ & trained at $L=16$ \\
\midrule
\multicolumn{7}{l}{\emph{Our eight-layer GCN, trained and tested at the same size}} \\
Eight-layer GCN, ours & -- & -- & ${-}1.089$ & ${-}0.930$ & ${-}0.987$ & 2 models, 25\,000 steps; 15\,000 at $L=256$ \\
\quad peak training memory (MB) & -- & -- & 4\,898 to 4\,916 & 4\,899 to 4\,917 & 12\,960 to 12\,979 & both models \\
\bottomrule
\end{tabular}
}
\caption{Every number on the benchmark at the harder coupling that we could place on one axis, in logMSE, lower better; the
text above describes each block. The runs at a quarter of the learning rate and at twice the steps are in \cref{tab:models,tab:baselines}, and the learning-rate check in \cref{tab:lr}. Entries marked $\dagger$ are size transfers, except the
entries marked $\ddagger$, the diagonal of the published transfer table, where the model was trained
and tested at that size, so the number is in-domain.}
\label{tab:stand}
\end{table}

\begin{table}[htbp]
\centering
\scriptsize
\setlength{\tabcolsep}{3pt}
\resizebox{\linewidth}{!}{\begin{tabular}{@{}llrrcccccrr@{}}
\toprule
Model & $L$ & width & parameters & \multicolumn{3}{c}{each model, test} & mean & s.d. & validation & hours \\
\midrule
Dilated CNN & 16 & 136 & 2\,020\,145 & ${-}4.915$ & ${-}5.091$ & ${-}4.883$ & ${-}4.963$ & $0.112$ & ${-}4.973$ & 1.0 \\
 & 32 & 116 & 1\,955\,529 & ${-}4.777$ & ${-}4.886$ & ${-}4.928$ & ${-}4.864$ & $0.078$ & ${-}4.868$ & 3.9 \\
 & 64 & 104 & 1\,962\,481 & ${-}4.282$ & ${-}4.420$ & ${-}4.289$ & ${-}4.330$ & $0.078$ & ${-}4.332$ & 2.5 \\
 & 128 & 96 & 2\,005\,057 & ${-}3.909$ & ${-}3.893$ & ${-}3.947$ & ${-}3.916$ & $0.028$ & ${-}3.918$ & 2.8 \\
 & 256 & 88 & 1\,964\,689 & ${-}3.491$ & ${-}3.393$ & ${-}3.450$ & ${-}3.445$ & $0.049$ & ${-}3.444$ & 5.4 \\
\midrule
Small dilated CNN & 16 & 52 & 296\,297 & ${-}4.899$ & ${-}4.953$ & ${-}4.877$ & ${-}4.910$ & $0.039$ & ${-}4.917$ & 0.3 \\
 & 32 & 44 & 282\,393 & ${-}4.971$ & ${-}5.008$ & ${-}4.781$ & ${-}4.920$ & $0.122$ & ${-}4.921$ & 1.2 \\
 & 64 & 40 & 291\,441 & ${-}4.677$ & ${-}4.591$ & ${-}4.532$ & ${-}4.600$ & $0.073$ & ${-}4.602$ & 0.8 \\
 & 128 & 36 & 283\,177 & ${-}4.346$ & ${-}3.833$ & ${-}4.172$ & ${-}4.117$ & $0.261$ & ${-}4.118$ & 0.9 \\
 & 256 & 36 & 330\,121 & ${-}3.595$ & ${-}3.924$ & ${-}3.718$ & ${-}3.746$ & $0.166$ & ${-}3.747$ & 1.7 \\
\midrule
FNO & 16 & 48 & 2\,375\,201 & ${-}4.392$ & ${-}4.569$ & ${-}4.534$ & ${-}4.498$ & $0.094$ & ${-}4.526$ & 0.1 \\
 & 32 & 32 & 2\,367\,937 & ${-}3.768$ & ${-}3.432$ & ${-}3.711$ & ${-}3.637$ & $0.180$ & ${-}3.701$ & 0.2 \\
 & 64 & 32 & 2\,367\,937 & ${-}1.350$ & ${-}1.338$ & ${-}1.455$ & ${-}1.381$ & $0.064$ & ${-}1.372$ & 0.1 \\
 & 128 & 32 & 2\,367\,937 & ${-}0.208$ & ${-}0.156$ & ${-}0.121$ & ${-}0.161$ & $0.044$ & ${-}0.163$ & 0.1 \\
 & 256 & 32 & 2\,367\,937 & $+0.481$ & $+0.477$ & $+0.525$ & $+0.494$ & $0.026$ & $+0.495$ & 0.1 \\
\midrule
Small FNO & 16 & 16 & 265\,569 & ${-}4.162$ & ${-}4.028$ & ${-}3.970$ & ${-}4.053$ & $0.099$ & ${-}4.071$ & 0.1 \\
 & 32 & 12 & 334\,217 & ${-}2.826$ & ${-}2.878$ & ${-}2.669$ & ${-}2.791$ & $0.109$ & ${-}2.901$ & 0.1 \\
 & 64 & 12 & 334\,217 & ${-}0.511$ & ${-}0.540$ & ${-}0.576$ & ${-}0.542$ & $0.033$ & ${-}0.545$ & 0.1 \\
 & 128 & 12 & 334\,217 & $+0.140$ & $+0.303$ & $+0.194$ & $+0.212$ & $0.083$ & $+0.212$ & 0.1 \\
 & 256 & 12 & 334\,217 & $+0.654$ & $+0.657$ & $+0.624$ & $+0.645$ & $0.019$ & $+0.645$ & 0.1 \\
\midrule
FNO, every mode & 32 & 22 & 1\,987\,605 & ${-}4.551$ & ${-}4.377$ & ${-}4.475$ & ${-}4.467$ & $0.087$ & ${-}4.462$ & 0.1 \\
FNO, grid coordinates & 32 & 32 & 2\,368\,001 & ${-}3.266$ & ${-}3.455$ & ${-}3.412$ & ${-}3.378$ & $0.099$ & ${-}3.470$ & 0.2 \\
\midrule
Dilated CNN, every dilation one & 32 & 116 & 1\,955\,529 & ${-}4.770$ & ${-}4.729$ & ${-}4.561$ & ${-}4.687$ & $0.111$ & ${-}4.689$ & 2.6 \\
 & 128 & 96 & 2\,005\,057 & ${-}2.879$ & ${-}2.885$ & ${-}2.878$ & ${-}2.881$ & $0.004$ & ${-}2.893$ & 1.8 \\
Dilated CNN, twice the steps & 128 & 96 & 2\,005\,057 & ${-}4.394$ & ${-}4.315$ & ${-}4.418$ & ${-}4.375$ & $0.054$ & ${-}4.376$ & 5.5 \\
 & 256 & 88 & 1\,964\,689 & ${-}4.000$ & ${-}3.901$ & ${-}4.083$ & ${-}3.995$ & $0.091$ & ${-}3.995$ & 10.4 \\
Dilated CNN, rate $5\times10^{-4}$ & 32 & 116 & 1\,955\,529 & ${-}5.141$ & ${-}5.150$ & ${-}5.180$ & ${-}5.157$ & $0.021$ & ${-}5.159$ & 3.6 \\
 & 128 & 96 & 2\,005\,057 & ${-}4.492$ & ${-}4.116$ & ${-}4.457$ & ${-}4.355$ & $0.208$ & ${-}4.356$ & 2.7 \\
FNO, every mode & 64 & 11 & 1\,984\,679 & ${-}3.904$ & ${-}3.845$ & ${-}3.921$ & ${-}3.890$ & $0.040$ & ${-}3.888$ & 0.1 \\
\midrule
\multicolumn{4}{l}{FNO mode floor, 12 modes per axis} & \multicolumn{7}{l}{$L=16$: every mode kept, $L=32$: ${-}1.085$, $L=64$: ${-}0.017$, $L=128$: $+0.507$, $L=256$: $+0.743$} \\
\bottomrule
\end{tabular}
}
\caption{Every baseline model on the official test split, in logMSE, lower is better: the three
models, their mean and sample standard deviation, the mean on the validation split, and the longest
training time of the three in hours on one NVIDIA T4 (for a run split in two, the total). The
rows below the operator with grid coordinates are the matched control without strides, the dilated
network at twice the steps and at a quarter of the learning rate, and the operator that keeps every
mode at a second size. The mode floor binds the linear part of the operator at \fnoModes{} modes per
axis.}
\label{tab:baselines}
\end{table}

\begin{table}[htbp]
\centering
\small
\begin{tabular}{@{}llcc@{}}
\toprule
Model & learning rate & validation & against the standard rate \\
\midrule
Dilated CNN & $2\times10^{-3}$ (standard) & ${-}4.778$ & -- \\
 & $10^{-3}$ & ${-}5.017$ & ${-}0.238$ \\
 & $5\times10^{-4}$ & ${-}5.143$ & ${-}0.365$ \\
\midrule
FNO & $2\times10^{-3}$ (standard) & ${-}3.644$ & -- \\
 & $10^{-3}$ & ${-}3.467$ & $+0.177$ \\
 & $5\times10^{-4}$ & ${-}2.910$ & $+0.733$ \\
\bottomrule
\end{tabular}

\caption{The learning-rate check at $L=32$, one model per rate, read on the validation split. The last
column is the validation logMSE at that rate minus the validation logMSE of the same seed at the
standard rate; positive is worse.}
\label{tab:lr}
\end{table}

\section{The second coupling}
\label{app:secondcoupling}

\Cref{tab:secondcoupling} gives the numbers behind \cref{sec:secondcoupling}. LRIM publishes every
size at the exponent \benchSigmaEasy{} with the same number of configurations and the same split as at
the harder coupling, and we used those official files. Each quantity of
\cref{tab:secondcoupling} is computed as at the harder coupling. The operator is fitted to the
training labels by the solve of \cref{app:proofs} and scored on the test split; the coupling itself,
scored the same way, reaches \secondExactJLsixteenRound{} to \secondExactJLtwofiftysixRound{} from the
smallest size to the largest. The two strided filters are fitted to the training labels by least
squares and read once on the test split, the larger one with its basis chosen on the validation split
as in \cref{sec:strides}. The floor and predicting zero are computed with the spin
spectrum of \polySpectrumGraphs{} training configurations, and the floor uses the coupling at this
exponent. The dilated network is the large one of \cref{sec:archs}, trained with the same budget,
recipe and seeds as at the harder coupling. The best published results are the benchmark's own, for
\secondPublishedModel{} at both sizes \citep{mathys2026lrim}, since later work reports the harder coupling only.

\begin{table}[htbp]
\centering
\small
\setlength{\tabcolsep}{3.5pt}
\resizebox{\linewidth}{!}{\begin{tabular}{lccccc}
\toprule
 & \multicolumn{5}{c}{nodes} \\
\cmidrule(lr){2-6}
 & 256 & 1\,024 & 4\,096 & 16\,384 & 65\,536 \\
\midrule
The operator, fitted to the training labels & ${-}29.26$ & ${-}28.74$ & ${-}27.78$ & ${-}27.47$ & ${-}26.97$ \\
The larger strided filter, fitted to the training labels & -- & ${-}8.76$ & ${-}9.41$ & ${-}9.20$ & ${-}9.15$ \\
The fifty-coefficient strided filter, fitted to the training labels & ${-}8.58$ & ${-}8.34$ & ${-}8.30$ & ${-}8.22$ & ${-}8.21$ \\
The floor for a filter polynomial in the adjacency & ${-}2.53$ & ${-}2.79$ & ${-}3.07$ & ${-}3.35$ & ${-}3.65$ \\
\midrule
A dilated CNN & ${-}5.36 \pm 0.08$ & ${-}5.47 \pm 0.02$ & ${-}5.08 \pm 0.09$ & ${-}4.96 \pm 0.06$ & ${-}4.43 \pm 0.06$ \\
Best published, trained and tested at the same size & ${-}5.30$ & ${-}5.13$ & -- & -- & -- \\
\midrule
Predicting zero at every node & $+1.22$ & $+1.20$ & $+1.19$ & $+1.16$ & $+1.16$ \\
\bottomrule
\end{tabular}
}
\caption{The benchmark at its easier coupling, exponent \benchSigmaEasy{}, in test logMSE, lower is
better. Each row is defined as in \cref{tab:compact}. At \benchNodesLow{} nodes the basis of the larger
strided filter spans every symmetric kernel, so that cell is empty. The dilated network is the mean
and standard deviation of three models. The best published results come from the benchmark's in-domain table.}
\label{tab:secondcoupling}
\end{table}

\section{Further closed forms and a wider readout}
\label{app:closedforms}

\subsection{The cost of leaving out one level of strides}

The gain of a level of strides added in turn depends on the order (\cref{tab:closed}, part e), so
\cref{sec:strides} quotes the cost of leaving one level out of the fifty-coefficient basis instead,
from the same fits to the coupling: at $L=32$ the
three levels of strides, finest first, cost \atrousDropCostLevelzeroLthirtytwo{},
\atrousDropCostLeveloneLthirtytwo{} and \atrousDropCostLeveltwoLthirtytwo{}, and at $L=64$ the four
cost \atrousDropCostLevelzeroLsixtyfour{}, \atrousDropCostLeveloneLsixtyfour{},
\atrousDropCostLeveltwoLsixtyfour{} and \atrousDropCostLevelthreeLsixtyfour{}; leaving out the fine
offsets costs \atrousDropFineCostLthirtytwo{} and \atrousDropFineCostLsixtyfour{}. Fitted to the
training labels and read on the test split, leaving out the top level costs
\atrousLabelDropTopCostLthirtytwo{} and \atrousLabelDropTopCostLsixtyfour{}.

\subsection{A wider readout}
\label{app:protocol}

Replacing the single gate $s_i$ of \cref{eq:readout} by the nine offsets
$a$ with $\lVert a \rVert_\infty \le 1$,
\[
\hat{y}_i \;=\; b \;+\; \sum_{a} s_{i+a} \cdot (K_a \circledast s)_i ,
\]
with a free Fourier multiplier $K_a$ for each offset, gives \stridedHeadNineOffsetParams{} parameters at
$L=32$. It contains the single-gate readout as the case in which only $K_0$ is nonzero. We trained it
at $L=32$ only, by gradient descent with the optimizer and schedule of the quadratic readout (\cref{sec:archs}) for \readoutStepsFree{}
steps, and it reaches \stridedHeadThirtyTwoNineOffsetRound{} on the official test split, against
\stridedHeadThirtyTwoFreeRound{} for a single-gate multiplier trained the same way beside it. The single-gate
value in \cref{tab:compact} comes from a separate run at all five sizes. The trained nine-gate form
keeps every kernel with $a \ne 0$ inside the nine-site window.

\section{What we expected, and what happened}
\label{app:registered}

For each experiment we wrote down, before it ran, what we expected and what would count against the
expectation. The tables give each expectation, the probability we gave it where we gave one, and what
happened, met or not. The probability is the one we wrote down before the runs it concerns, as our
estimate that the expectation would hold. On the closed forms (\cref{tab:registered}) the
learned readout met its expectation at every size, and the band we expected for the best polynomial
filter held at some sizes and degrees and failed at others. The floor of \cref{prop:floor}, computed
with the spin spectrum of the training configurations, replaced that band.

\Cref{tab:registeredbaselines} covers the standard baselines. \basePredSentence{} Among those not met,
the small dilated network was not worse than the large one, the Fourier neural operator fell below the
band we had set around the floor of its truncated linear class at every size but the largest, and keeping every mode
improved the operator by less than we expected. The learning-rate check counts as not met because it held for the operator
but not for the dilated network, which improved at the lower rates.

\Cref{tab:registeredlater} holds the \baseLaterPredCountWord{} expectations for four additional
experiments: the dilated network with every dilation set to one, at twice the steps at the two largest
sizes, and at a quarter of the learning rate, and the operator that keeps every mode at a second size.
\baseLaterPredSentence{} Twice the steps left most of the loss with size in place at the largest size,
as we expected, and at \latticeLonetwentyeightNodes{} nodes recovered slightly more than we allowed
for. The first three blocks of \cref{tab:registeredgraph} hold the \baseGraphPredCountWord{} expectations for the graph networks of
\cref{sec:graphnets}, the longer recipe for two of them, and two more models of each at that
recipe. \baseGraphPredSentence{} The gated network
did not come close to its published value under our protocol, the separate weights per edge direction
improved it less than we expected, the network with strides did not come near the dilated network, and global
attention added more than we expected. At the longer recipe the gated network came close to its
published value and both networks stayed above the dilated network by at least the margin we had
set, but the lead of the network with strides over the gated network shrank below the margin we had
set for it. With two more models of each, both results held, as we expected. Its last block holds
the \gatedDcnnPredCountWord{} expectations for the dilated network with the target's gate, and
\gatedDcnnPredMetWord{} was met: the gate improved the network by \gatedDcnnGainLsixteen{} at
\benchNodesLow{} nodes and \gatedDcnnGainLthirtytwo{} at \latticeLthirtytwoNodes{} nodes.

\Cref{tab:registeredjournal} holds the expectations for the polynomial spectral networks of
\cref{sec:spectral} and for the easier coupling of \cref{sec:secondcoupling}, each written down
before any run on the official data. All of them were met.

\begin{table}[htbp]
\centering
\scriptsize
\setlength{\tabcolsep}{3pt}
\begin{tabular}{>{\raggedright\arraybackslash}p{0.18\linewidth}>{\raggedright\arraybackslash}p{0.25\linewidth}>{\raggedright\arraybackslash}p{0.18\linewidth}>{\raggedright\arraybackslash}p{0.34\linewidth}}
\toprule
Experiment & What we expected & What would have counted against it & What happened \\
\midrule
The learned quadratic readout, at all five sizes & $-6.5$ or better with the larger strided basis at every size, $-11$ or better with the free multiplier wherever it fits in memory & a size where either target is missed & met at every size; the free multiplier also fits at $L=256$ \\
\addlinespace[2pt]
The best polynomial filter in the adjacency, at four sizes & within 0.3 of $-1.9$ at every size & a size outside that band & partly met: at degree 128 it holds at $L=64$ and $L=128$ and fails at $L=16$ and $L=32$; at degree 64 it holds at $L=32$, 64 and 128 and fails at $L=16$. The floor of Proposition 1, exact for the training spectrum, replaces the band \\
\bottomrule
\end{tabular}

\caption{What we expected from each experiment on the closed forms, what would have counted against
it, and what happened.}
\label{tab:registered}
\end{table}

\begin{table}[htbp]
\centering
\scriptsize
\setlength{\tabcolsep}{3pt}
\begin{tabular}{@{}>{\raggedright\arraybackslash}p{0.40\linewidth}c>{\raggedright\arraybackslash}p{0.47\linewidth}@{}}
\toprule
What we expected & probability & What happened \\
\midrule
The dilated CNN passes the polynomial floor at every size & 75\,\% & not met at $L=256$, where it scores ${-}3.445$ against a floor of ${-}3.518$; met at the other sizes \\
\addlinespace[1pt]
The dilated CNN lands between $-5.3$ and $-3.3$ at $L=32$ & 70\,\% & met: ${-}4.864$ \\
\addlinespace[1pt]
The dilated CNN beats the best published in-domain result at $L=16$ and $L=32$ & 45\,\% & met: ${-}4.963$ against ${-}4.345$ (the benchmark's transfer table, trained and tested at $L=16$) and ${-}4.864$ against ${-}4.134$; against the stronger results published since, ${-}4.963$ against ${-}4.701$ and ${-}4.864$ against ${-}4.852$, ahead by $0.26$ and $0.01$ (the second within the spread of seeds) \\
\addlinespace[1pt]
The dilated CNN is at least $1.5$ below our eight-layer baseline at $L=64$, 128 and 256 & 75\,\% & met: by $3.24$, $2.99$, $2.46$ \\
\addlinespace[1pt]
The dilated CNN stays at least $1.5$ above the larger strided filter from $L=32$ on & 95\,\% & met: $1.89$ to $4.14$ above \\
\addlinespace[1pt]
The small dilated CNN is worse than the dilated CNN at $L=32$ by $0.1$ to $1.2$ & 60\,\% & not met: it is $0.056$ better \\
\addlinespace[1pt]
The FNO lies between $0.5$ below and $0.3$ above its mode floor at $L=32$ to 256 & 55\,\% & not met at $L=32$, 64 and 128, where it lies $2.55$, $1.36$, $0.67$ below the floor; met at $L=256$ ($0.25$ below) \\
\addlinespace[1pt]
The FNO is at least $1.5$ worse than the dilated CNN at $L=64$, 128 and 256 & 80\,\% & met: by $2.95$, $3.75$, $3.94$ \\
\addlinespace[1pt]
The FNO at $L=16$, where it keeps every mode, passes the polynomial floor & 70\,\% & met: ${-}4.498$ against ${-}2.475$ \\
\addlinespace[1pt]
The FNO at $L=128$ is worse than our eight-layer baseline & 60\,\% & met: ${-}0.161$ against ${-}0.930$ \\
\addlinespace[1pt]
The small FNO lies within $0.3$ of the FNO at $L=64$, 128 and 256 & 65\,\% & not met at $L=64$ and 128, where the small one is $0.84$ and $0.37$ worse; met at $L=256$ ($0.15$) \\
\addlinespace[1pt]
Keeping every mode at $L=32$ improves the FNO by at least $1.0$ & 70\,\% & not met: it improves by $0.83$, to ${-}4.467$ \\
\addlinespace[1pt]
Grid coordinates move the FNO at $L=32$ by less than $0.2$ & 70\,\% & not met: with them it is $0.26$ worse, at ${-}3.378$ \\
\addlinespace[1pt]
No learning rate of the check beats the rate of the main runs by more than $0.2$ on validation & 65\,\% & the FNO: met, the lower rates are $0.18$ and $0.73$ worse; the dilated CNN: not met, the lower rates are $0.24$ and $0.36$ better \\
\bottomrule
\end{tabular}

\caption{What we expected of the standard baselines, the probability we gave each expectation before
the runs it concerns, and what happened on the official test split, except for the learning-rate
check, which is read on validation. The outcome of the third expectation is scored against the best
published results at the time of registration, and then against the stronger results of
\citet{carotti2026ghr} published since.}
\label{tab:registeredbaselines}
\end{table}

\begin{table}[htbp]
\centering
\scriptsize
\setlength{\tabcolsep}{3pt}
\begin{tabular}{@{}>{\raggedright\arraybackslash}p{0.40\linewidth}c>{\raggedright\arraybackslash}p{0.47\linewidth}@{}}
\toprule
What we expected & probability & What happened \\
\midrule
\multicolumn{3}{@{}l}{\emph{The same network with every dilation one, at $L=32$ and $L=128$}} \\
The same network with every dilation one is worse than the dilated CNN by at least $0.3$ at $L=32$ & 65\,\% & not met: ${-}4.687$ against ${-}4.864$, $0.18$ worse \\
\addlinespace[1pt]
The same network with every dilation one is worse than the dilated CNN by at least $0.5$ at $L=128$ & 75\,\% & met: ${-}2.881$ against ${-}3.916$, $1.04$ worse \\
\addlinespace[1pt]
Its shortfall against the dilated CNN is larger at $L=128$ than at $L=32$ & 70\,\% & met: it is $1.04$ worse at $L=128$ and $0.18$ worse at $L=32$ \\
\addlinespace[1pt]
At $L=128$ it lies no more than $0.5$ below the reference for a filter of its reach & 65\,\% & met: ${-}2.881$ against a reference of ${-}2.955$ \\
\addlinespace[1pt]
At $L=32$ it passes the polynomial floor & 80\,\% & met: ${-}4.687$ against ${-}2.678$ \\
\midrule
\multicolumn{3}{@{}l}{\emph{Twice the steps for the dilated CNN, and every mode kept at a second size}} \\
The dilated CNN at twice the steps is at least $0.1$ better than at the standard budget at $L=128$ & 70\,\% & met: ${-}4.375$ against ${-}3.916$ \\
\addlinespace[1pt]
The dilated CNN at twice the steps is at least $0.1$ better than at the standard budget at $L=256$ & 65\,\% & met: ${-}3.995$ against ${-}3.445$ \\
\addlinespace[1pt]
The dilated CNN at twice the steps passes the polynomial floor at $L=256$ & 60\,\% & met: ${-}3.995$ against ${-}3.518$ \\
\addlinespace[1pt]
The dilated CNN at twice the steps at $L=128$ stays at least $0.5$ above the dilated CNN at $L=32$ & 80\,\% & not met: ${-}4.375$ against ${-}4.864$ \\
\addlinespace[1pt]
The dilated CNN at twice the steps at $L=256$ stays at least $0.8$ above the dilated CNN at $L=32$ & 85\,\% & met: ${-}3.995$ against ${-}4.864$ \\
\addlinespace[1pt]
The FNO that keeps every mode is at least $1.5$ better than the FNO at $L=64$ & 70\,\% & met: ${-}3.890$ against ${-}1.381$ \\
\addlinespace[1pt]
The FNO that keeps every mode is still worse than the dilated CNN at $L=64$ & 75\,\% & met: ${-}3.890$ against ${-}4.330$ \\
\addlinespace[1pt]
The FNO that keeps every mode passes the polynomial floor at $L=64$ & 65\,\% & met: ${-}3.890$ against ${-}2.949$ \\
\midrule
\multicolumn{3}{@{}l}{\emph{A quarter of the learning rate for the dilated CNN, at $L=32$ and $L=128$}} \\
The dilated CNN at a quarter of the learning rate is at least $0.15$ better than at the standard rate at $L=32$ & 75\,\% & met: ${-}5.157$ against ${-}4.864$ \\
\addlinespace[1pt]
The dilated CNN at a quarter of the learning rate is at least $0.15$ better than at the standard rate at $L=128$ & 60\,\% & met: ${-}4.355$ against ${-}3.916$ \\
\addlinespace[1pt]
Its gain at $L=128$ is no more than twice its gain at $L=32$ & 70\,\% & met: a gain of $0.44$ at $L=128$ against $0.29$ at $L=32$ \\
\bottomrule
\end{tabular}

\caption{What we expected of the additional experiments, the probability we gave each expectation before
the runs it concerns, and what happened on the official test split. The reference for a filter of its
reach, in the first block, is the best readout of the target's form whose filter is confined to the
square the network without dilations can reach; it is a reference, not a bound on a nonlinear network.}
\label{tab:registeredlater}
\end{table}

\begin{table}[htbp]
\centering
\scriptsize
\setlength{\tabcolsep}{3pt}
\begin{tabular}{@{}>{\raggedright\arraybackslash}p{0.40\linewidth}c>{\raggedright\arraybackslash}p{0.47\linewidth}@{}}
\toprule
What we expected & probability & What happened \\
\midrule
\multicolumn{3}{@{}l}{\emph{Four graph networks under the same protocol, at $L=16$ and $L=32$}} \\
GatedGCN at $L=16$ lands within $0.5$ of its published value & 60\,\% & not met: ${-}2.951$ against ${-}3.844$, $0.89$ above \\
\addlinespace[1pt]
The per-direction GatedGCN is at least $0.3$ better than GatedGCN at $L=16$ & 65\,\% & not met: ${-}3.022$ against ${-}2.951$, $0.07$ better \\
\addlinespace[1pt]
The per-direction GatedGCN is at least $0.3$ better than GatedGCN at $L=32$ & 60\,\% & not met: ${-}3.283$ against ${-}3.249$, $0.03$ better \\
\addlinespace[1pt]
The strided GatedGCN lies within $0.3$ of the dilated CNN at $L=16$ and at $L=32$ & 50\,\% & not met: at $L=16$, ${-}3.456$ against ${-}4.963$, $1.51$ above; at $L=32$, ${-}3.363$ against ${-}4.864$, $1.50$ above \\
\addlinespace[1pt]
GPS with RWSE is not more than $0.3$ better than GatedGCN at $L=16$ & 60\,\% & not met: ${-}3.451$ against ${-}2.951$, $0.50$ better \\
\addlinespace[1pt]
All four pass the polynomial floor at $L=16$ and at $L=32$ & 85\,\% & met: at $L=16$, the weakest at ${-}2.951$ against a floor of ${-}2.475$; at $L=32$, where GPS with RWSE was not tested, the weakest at ${-}3.249$ against a floor of ${-}2.678$ \\
\midrule
\multicolumn{3}{@{}l}{\emph{Two graph networks at four times the steps and half the learning rate, at $L=16$}} \\
GatedGCN at the longer recipe comes within $0.4$ of its published value or better & 55\,\% & met: ${-}3.747$ against ${-}3.844$ \\
\addlinespace[1pt]
GatedGCN at the longer recipe is at least $0.3$ better than the same seed at the standard budget & 75\,\% & met: ${-}3.747$ against ${-}2.906$ \\
\addlinespace[1pt]
The strided GatedGCN at the longer recipe is at least $0.3$ better than GatedGCN at the longer recipe & 65\,\% & not met: ${-}3.847$ against ${-}3.747$, $0.10$ better \\
\addlinespace[1pt]
Both stay at least $0.5$ above the dilated CNN at $L=16$ & 80\,\% & met: ${-}3.747$ and ${-}3.847$ against ${-}4.963$ \\
\midrule
\multicolumn{3}{@{}l}{\emph{Two more models of each at the longer recipe, at $L=16$}} \\
Over three seeds, the strided GatedGCN's lead over GatedGCN at the longer recipe is below $0.3$ & 70\,\% & met: ${-}3.985$ against ${-}3.819$, $0.17$ better \\
\addlinespace[1pt]
Both stay at least $0.8$ above the dilated CNN at $L=16$, three-seed means & 85\,\% & met: ${-}3.819$ and ${-}3.985$ against ${-}4.963$ \\
\midrule
\multicolumn{3}{@{}l}{\emph{The dilated CNN with the target's gate, at $L=16$ and $L=32$}} \\
The gated dilated CNN is at least $0.3$ better than the dilated CNN at $L=16$ and at $L=32$ & 75\,\% & not met: at $L=16$, ${-}5.017$ against ${-}4.963$, $0.05$ better; at $L=32$, ${-}5.095$ against ${-}4.864$, $0.23$ better \\
\addlinespace[1pt]
The gated dilated CNN is at least $1.0$ better than the dilated CNN at $L=16$ and at $L=32$ & 55\,\% & not met: at $L=16$, ${-}5.017$ against ${-}4.963$, $0.05$ better; at $L=32$, ${-}5.095$ against ${-}4.864$, $0.23$ better \\
\addlinespace[1pt]
The gated dilated CNN reaches $-6.5$ or lower at $L=32$, the strided filter's level & 25\,\% & not met: ${-}5.095$ \\
\bottomrule
\end{tabular}

\caption{What we expected of the graph networks and of the gated dilated CNN, the probability we gave
each expectation before the runs it concerns, and what happened on the official test split. The
per-direction GatedGCN has a separate weight for each edge direction, and the strided GatedGCN is
that network with edges at one more stride in each layer, cycling through $2, 4, \ldots, L/2$. The
expectations at four times the steps were written for one model of each and are read on that model;
the third block reads the means of three. The gated dilated CNN multiplies the dilated CNN's output
by the node's spin and adds a learned constant; one model at each size.}
\label{tab:registeredgraph}
\end{table}

\begin{table}[htbp]
\centering
\scriptsize
\setlength{\tabcolsep}{3pt}
\begin{tabular}{@{}>{\raggedright\arraybackslash}p{0.40\linewidth}c>{\raggedright\arraybackslash}p{0.47\linewidth}@{}}
\toprule
What we expected & probability & What happened \\
\midrule
\multicolumn{3}{@{}l}{\emph{Two polynomial spectral networks, at $L=16$, $32$ and $64$}} \\
The gated linear filter lies within $0.05$ of the best polynomial of its degree at every size, for $K=8$ and for $K=L$ & 90\,\% & met: within $0.005$ everywhere \\
\addlinespace[1pt]
SIGN at $K=L$ does not beat the floor by more than $0.1$ at any size & 65\,\% & met: $0.42$, $0.70$, $1.00$ above the floor \\
\addlinespace[1pt]
SIGN at $K=L$ lies within $0.3$ of the best polynomial of degree $L$ at every size & 50\,\% & met: $0.08$, $0.06$, $0.05$ below it \\
\addlinespace[1pt]
SIGN at $K=8$ lies within $0.3$ of the best polynomial of degree eight at every size & 55\,\% & met: $0.06$, $0.03$, $0.03$ below it \\
\midrule
\multicolumn{3}{@{}l}{\emph{The easier coupling, exponent $1.5$, at every size}} \\
Least squares on the training labels recovers the operator to ${-}20$ or lower at every size & 90\,\% & met: ${-}27.0$ to ${-}29.3$ \\
\addlinespace[1pt]
The polynomial floor lies at least $2.0$ above the fifty-coefficient filter fitted to the labels at $L=32$ and $L=64$ & 85\,\% & met: $5.5$ and $5.2$ above \\
\addlinespace[1pt]
The fifty-coefficient filter fitted to the labels is at least $1.0$ below the best published network at $L=32$ & 70\,\% & met: ${-}8.34$ against ${-}5.13$ \\
\addlinespace[1pt]
The same at $L=16$ & 60\,\% & met: ${-}8.58$ against ${-}5.30$ \\
\addlinespace[1pt]
The dilated CNN is below the best published network at $L=16$ and at $L=32$ & 55\,\% & met: ${-}5.364$ against ${-}5.296$ and ${-}5.472$ against ${-}5.134$ \\
\addlinespace[1pt]
The dilated CNN stays at least $1.0$ above the fifty-coefficient filter at every size & 85\,\% & met: $2.9$ to $3.8$ above \\
\bottomrule
\end{tabular}

\caption{What we expected of the polynomial spectral networks and at the easier coupling, the
probability we gave each expectation before the runs it concerns, and what happened on the official
test split. $K$ is the number of rounds of propagation. The fifty-coefficient filter is the strided
filter of \cref{sec:strides}, fitted to the training labels at the easier coupling; the dilated CNN
entries are means of three models.}
\label{tab:registeredjournal}
\end{table}

\end{document}